\documentclass[preprint,12pt,nopreprintline]{elsarticle}
\usepackage[a4paper,margin=1in]{geometry}
\usepackage{amsmath,amssymb,amsfonts,amsthm,mathtools}
\usepackage{bm}
\usepackage{booktabs}
\usepackage{multirow}
\usepackage{graphicx}
\usepackage{algorithm}
\usepackage{algpseudocode}
\usepackage{enumitem}
\usepackage[section]{placeins}
\usepackage{needspace}
\usepackage{xcolor}
\usepackage{hyperref}
\usepackage[nameinlink,capitalize]{cleveref}
\crefname{assumption}{Assumption}{Assumptions}
\crefname{proposition}{Proposition}{Propositions}
\Crefname{proposition}{Proposition}{Propositions}
\crefname{remark}{Remark}{Remarks}
\Crefname{remark}{Remark}{Remarks}
\hypersetup{hidelinks,pdftitle={Latent Annealed Posterior Sampling with a Variance-Preserving Bridge for Sparsely Observed Random Fields},pdfauthor={Jiankun Lei, Tian Qi}}

\usepackage{threeparttable}
\usepackage{makecell}
\usepackage{pifont}
\journal{}
\date{October 14, 2026}
\makeatletter
\let\laps@pprintTitle\ps@pprintTitle
\def\ps@pprintTitle{\laps@pprintTitle\let\laps@oddfoot\@oddfoot\def\@oddfoot{\laps@oddfoot\unskip}\let\@evenfoot\@oddfoot}
\makeatother

\numberwithin{equation}{section}

\newtheorem{theorem}{Theorem}[section]

\newtheorem{proposition}[theorem]{Proposition}

\theoremstyle{remark}
\newtheorem{remark}[theorem]{Remark}

\newcommand{\R}{\mathbb{R}}
\newcommand{\X}{\mathcal{X}}
\newcommand{\U}{\mathcal{U}}
\newcommand{\A}{\mathcal{A}}
\newcommand{\N}{\mathcal{N}}
\newcommand{\E}{\mathbb{E}}
\newcommand{\appendixsection}[1]{%
  \FloatBarrier
  \Needspace{10\baselineskip}%
  \refstepcounter{section}%
  \setcounter{subsection}{0}%
  \setcounter{table}{0}%
  \setcounter{figure}{0}%
  \section*{Appendix~\thesection\quad #1}%
  \addcontentsline{toc}{section}{\protect\numberline{Appendix~\thesection}#1}%
}
\newcommand{\appendixsubsection}[1]{%
  \Needspace{7\baselineskip}%
  \refstepcounter{subsection}%
  \subsection*{\thesubsection\quad #1}%
  \addcontentsline{toc}{subsection}{\protect\numberline{\thesubsection}#1}%
}
\newcommand{\vz}{\mathbf{z}}
\newcommand{\vx}{\mathbf{x}}
\newcommand{\vy}{\mathbf{y}}
\newcommand{\vu}{\mathbf{u}}

\newcommand{\veps}{\boldsymbol{\epsilon}} 
\newcommand{\vzero}{\mathbf{0}}

\begin{document}

\begin{frontmatter}
\title{Latent Annealed Posterior Sampling with a Variance-Preserving Bridge
for Sparsely Observed Random Fields}

\author[inst1]{Jiankun Lei\corref{cor1}}
\ead{love2019@cxtc.edu.cn}
\author[inst1]{Tian Qi}

\cortext[cor1]{Corresponding author}

\affiliation[inst1]{
  organization={School of Statistics and Mathematics,
  Yunnan University of Finance and Economics},
  addressline={No. 237 Longquan Road, Wuhua District},
  city={Kunming},
  postcode={650221},
  state={Yunnan},
  country={China}
}
\begin{abstract}
Posterior inference for random fields from sparse observations presents two major challenges: accurately characterizing complex spatial uncertainty and preventing conditional sampling costs from growing substantially as the spatial grid is refined. To address these challenges, we propose latent annealed posterior sampling (LAPS). LAPS first maps the random field into a latent space with reduced dimensionality, while a coordinate decoder composed of Branch and Trunk networks enables field evaluation at arbitrary spatial locations. Flow matching is then used to learn the latent prior distribution. Based on this learned prior, posterior sampling is performed entirely in the latent space, with the observation likelihood incorporated progressively through an annealing procedure. To control local covariance inflation, we introduce a variance-preserving (VP) bridge that allocates a covariance budget between the transported latent state and a Gaussian perturbation. Numerical experiments demonstrate that LAPS accurately approximates posterior distributions, preserves distinct posterior modes, and reduces the dependence of conditional sampling costs on spatial resolution. Across two Gaussian benchmarks, LAPS attains small errors against exact posterior references, with sliced Wasserstein distances only 2 to 3 percent higher than the reference values obtained from exact posterior samples. In a bimodal posterior problem, LAPS reduces the sliced Wasserstein distance from 0.2693 to 0.1499 and consistently recovers both modes across all test scenarios. For a two dimensional Navier–Stokes random field, LAPS substantially improves marginal posterior calibration, while its evaluation time remains nearly constant as the output grid is refined from 64² to 512². These results demonstrate that LAPS achieves a favorable tradeoff among posterior accuracy, uncertainty calibration, and computational efficiency.
\end{abstract}


\end{frontmatter}

\section{Introduction}
\label{sec:introduction}
Reconstructing a spatial field from a few noisy point measurements is a recurring task in science and engineering. Because sparse data are consistent with many fields, the quantity of interest is not a single reconstruction but the posterior distribution of the field given the data. The Bayesian formulation of such inverse problems is well established, both for discretized unknowns and directly on function spaces \citep{kaipio2005statistical,stuart2010inverse}. In practice, two requirements compete: the prior must be expressive enough to describe realistic fields, and the posterior must remain affordable to sample when one prior is reused across many observation sets and output resolutions.

Gaussian-process priors are the classical choice. Under a linear Gaussian observation model, their posterior is available in closed form \citep{rasmussen2006gpml}. For nonlinear likelihoods, Markov chain Monte Carlo (MCMC) methods formulated on function space remain efficient under mesh refinement \citep{cotter2013mcmc}, exploit the low-dimensional subspace informed by the data \citep{cui2016dimension} and the local geometry of the posterior \citep{beskos2017geometric}, and extend to certain non-Gaussian priors \citep{vollmer2015dimension}. These classical formulations rely on tractable reference measures or specified prior structure; an implicit prior represented by a learned flow requires an additional conditioning construction. Fields with multimodal or otherwise strongly non-Gaussian laws, such as those generated by nonlinear dynamics, cannot be described by a covariance function alone, which motivates priors learned from data.

Learned stochastic-process models provide this expressiveness. Neural processes learn a map from context observations to predictive distributions \citep{garnelo2018cnp,kim2019anp}, and diffusion and flow-matching models have been formulated directly on function spaces \citep{lim2025ddo,kerrigan2024ffm}, where neural operators make them independent of the discretization \citep{kovachki2023neuraloperator,li2021fno}. Such a prior can be sampled on any grid, but it is defined only implicitly, through a sampler or a learned velocity field, so conditioning it on new observations is a separate problem.

Two strategies address this problem. Amortized methods learn the conditional distribution itself \citep{abuhamad2025flownp,baldassari2023conditional,moss2025fnope}, which fixes the observation model at training time. Likelihood-guided methods instead condition a reusable prior at test time. For diffusion priors in finite dimensions, posterior sampling steers the reverse process with the gradient of the likelihood \citep{chung2023dps}, and this approach has been extended to function spaces \citep{yao2025fundps}. Operator flow matching conditions through learned finite-dimensional densities \citep{shi2025ofm}, whereas Flow Annealing Posterior Sampling (FAPS) conditions a pretrained function-space flow prior without evaluating its density \citep{shi2026faps}.

The cost of test-time conditioning is tied to the output grid whenever the Markov state and its covariance operations are kept on an explicit field discretization, as in the released FAPS implementation. Each update is repeated thousands of times per scenario, so the cost grows with the requested output resolution even when the observations are sparse. This is a limitation of the implementation class, not a lower bound for function-space samplers. A natural remedy is to carry out the repeated conditioning in a fixed-dimensional latent space and to reconstruct the field only once, at the end. Latent spaces already serve this purpose elsewhere: running generation in the latent space of an autoencoder is what makes high-resolution diffusion models affordable \citep{rombach2022ldm}, operator learning gains accuracy and efficiency in a learned latent space \citep{kontolati2024latent}, and deep generative models are established tools for inverse problems \citep{patel2022deep,goh2022vae,whang2021composing,song2024resample}. The price is that the posterior is confined to what the generator can represent: representation error and dataset bias limit the solution \citep{asim2020invertible}, and errors in a learned prior carry over to the posterior \citep{hosseini2026error}. For random fields, two questions therefore remain open: whether the separation of conditioning from the output grid can be specified precisely enough to predict a measurable cost, and how faithful the resulting posterior is when assessed against exact references rather than against other approximations.

To address these questions, we propose latent annealed posterior sampling (LAPS). A coordinate-based autoencoder with a Branch--Trunk decoder \citep{lu2021deeponet} represents each field by a fixed-dimensional latent code, and flow matching learns the prior over these codes \citep{lipman2023flow,tong2024cfm}. The transport--\allowbreak update--\allowbreak re-bridging cycle of FAPS then runs entirely in latent space: the likelihood evaluates the decoder only at the observation sites, and the output grid is evaluated once, after sampling. LAPS thus inherits the annealing principle of FAPS, and its central computational contribution is to separate repeated posterior conditioning from final field evaluation. The decoder also defines a single coherent field-level prior and posterior across coordinate sets, although this consistency alone does not guarantee that the learned law matches the original field distribution.

The annealing construction shared by the two methods raises a second issue. When the anchor of a local update already carries the prior covariance, the original local Gaussian kernel adds variance without reducing the contribution of the anchor. LAPS therefore uses a variance-preserving (VP) bridge, which allocates a single covariance budget, \(\alpha_k^2+\lambda_k^2=1\), and fixes the two coefficients through the endpoint-to-noise ratio of the implemented path. For a general latent prior, the resulting covariance error separates into a mismatch of the transported anchor and a mismatch between the encoded and the generated covariance. This identity is local. The inherited annealing guarantee requires exact conditional draws, whereas LAPS uses approximate local targets and a finite number of Metropolis-adjusted Langevin (MALA) updates, and a matched covariance preserves neither non-Gaussian higher-order structure nor the invariance of the full recursion.

The experiments are organized around these distinctions. Exact Gaussian and mixture posteriors test distributional accuracy and the retention of two plausible field configurations, while Navier--Stokes fields test predictive accuracy and marginal calibration. Relative to FAPS, LAPS yields smaller errors against the exact references and better marginal calibration on Navier--Stokes, whereas FAPS attains a lower continuous ranked probability score (CRPS) and field error there. A resolution sweep assesses the dependence on final grid size within each implementation; cross-method timing ratios remain exploratory because the recorded devices and timing protocols are not fully matched. Likelihood-free checks examine local covariance preservation and the departure of the full recursion from invariance, and ablations paired over five training seeds separate the effect of the bridge coefficients from that of the annealing schedule. Direct MCMC under the same learned prior is comparably accurate on the one-dimensional benchmarks at a matched budget and less well calibrated on Navier--Stokes at the tested budget, so the experiments do not establish a general advantage of annealing. The main contribution is therefore a computationally separated latent posterior sampler for random fields; the VP bridge is a local covariance-control mechanism within that sampler, not a guarantee of uniformly improved accuracy or of full-recursion invariance. In summary, our contributions are as follows:
\begin{enumerate}
    \item We perform repeated annealed updates in a fixed-dimensional latent space, thereby decoupling posterior conditioning from the output resolution and limiting the growth of online sampling costs as the spatial resolution increases.
    \item We introduce a variance preserving bridge that prevents excess prior variance from accumulating during annealing. By maintaining a consistent variance budget across annealing levels, this approach improves posterior fidelity and yields uncertainty estimates that are substantially better calibrated against the reference posterior.
    \item We conduct numerical experiments on Gaussian, bimodal, and two dimensional Navier–Stokes random fields. The results show that LAPS accurately approximates posterior distributions, preserves multimodal structures, improves uncertainty calibration, and substantially reduces computational costs.
\end{enumerate}
\section{Related Work}

This section compares LAPS with the most closely related methods along the three design choices that define it: when the prior is conditioned, how the conditional distribution is sampled, and where the sampling state resides.

\subsection{Amortized and Test-Time Conditioning}

Amortized models learn the conditional distribution directly. Conditional and attentive neural processes (CNP and ANP) map a context set to a predictive distribution \citep{garnelo2018cnp,kim2019anp}, neural diffusion processes and flow-matching neural processes (FlowNP) model more expressive conditionals \citep{dutordoir2023ndp,abuhamad2025flownp}, and conditional score models and simulation-based inference with neural operators amortize over observations of a prescribed forward model \citep{baldassari2023conditional,moss2025fnope}. These models predict at arbitrary target coordinates, but their conditioning is tied to the context--target distribution or forward model seen during training, so a new observation operator or likelihood falls outside their scope. CNP, ANP and FlowNP serve as the amortized baselines in \Cref{sec:experiments}.

Test-time methods keep an unconditional prior and introduce the likelihood only at inference; they differ in what they require of that prior. Neural operator flows and operator flow matching (OFM) provide finite-dimensional marginal densities at queried coordinates \citep{shi2024opflow,shi2025ofm}, which permit direct Bayesian updates, for instance by stochastic-gradient Langevin dynamics \citep{shi2025ofm}, at the cost of invertible transport and Jacobian-trace estimation. Priors that are available only through sampling, such as functional flow matching and function-space diffusion \citep{kerrigan2024ffm,lim2025ddo}, require a separate conditioning mechanism: guided diffusion steers a score-based prior with observation gradients \citep{yao2025fundps}, and FAPS conditions a frozen flow-matching prior without evaluating its density \citep{shi2026faps}. LAPS belongs to this last group, but its flow prior is defined on latent codes rather than on fields.

\subsection{Annealing and Markov Kernels}

FAPS is the method closest to LAPS, which adopts its level structure \citep{shi2026faps}. At each annealing level, FAPS transports a noisy state to the clean endpoint of the frozen flow, applies a likelihood-guided Langevin update with a low-rank covariance preconditioner, and re-bridges the result to the next level; one prior can thereby serve different masks, noise levels and spatial grids. Annealing through a sequence of intermediate distributions has a long history in Monte Carlo \citep{neal2001ais,delmoral2006smc}; here, the sequence follows the noise path of the flow prior instead of tempering the likelihood. Because the two methods share this construction, the variance analysis of \Cref{sec:analysis-bridge} applies to both. The baseline is run at its published configuration (\Cref{sec:experimental-design}).

Within each level, LAPS uses a Metropolis-adjusted Langevin kernel \citep{roberts1996exponential}. Its preconditioner follows likelihood-informed and geometric samplers, which shape proposals according to the local geometry of the target \citep{girolami2011riemann,cui2016dimension,beskos2017geometric}; a Gauss--Newton matrix held fixed within each level keeps the Metropolis correction exact for the local target (\Cref{app:mala}). Function-space MCMC also offers an alternative to annealing altogether. The preconditioned Crank--Nicolson (pCN) proposal is reversible with respect to a Gaussian reference measure, and its acceptance rate does not degenerate with dimension \citep{stuart2010inverse,cotter2013mcmc}. A flow prior supplies such a reference, because its posterior can be pulled back to the Gaussian base distribution. We use this direct sampler as a control (\Cref{app:direct-mcmc}).

\subsection{Latent Representations and Representation Error}

When an implementation keeps an explicitly discretized field state, every repeated gradient and covariance operation scales with the number of represented field values. This statement describes that class of implementations (\Cref{sec:analysis-cost}); it is not a lower bound for function-space algorithms in general. A latent state removes this dependence but introduces a representation. Deep generative models enter inverse problems in several ways: generative adversarial network (GAN) priors enable MCMC in a low-dimensional latent space \citep{patel2022deep}, variational autoencoders are trained to output an approximate posterior directly \citep{goh2022vae}, normalizing-flow approaches compose a pretrained prior with a second flow that approximates the conditional \citep{whang2021composing}, and latent diffusion solvers enforce data consistency through the decoder and return the estimate to the noisy latent manifold \citep{song2024resample}. These methods reduce the effective dimension of the problem, but not the approximation error of the representation. Representation error and dataset bias pass into the solution, an effect that invertible generators mitigate \citep{asim2020invertible}, and, under suitable assumptions, the posterior inherits in Wasserstein-1 distance the error rate of a prior learned as a minimum Wasserstein-2 generative model \citep{hosseini2026error}.

Coordinate-based operator networks represent fields at arbitrary locations \citep{lu2021deeponet,kovachki2023neuraloperator}, and learning them in a latent space reduces their cost \citep{kontolati2024latent}. Latent conditioning is therefore not new in itself. LAPS combines it with a coordinate-decoded random-field model and annealed likelihood updates, states the resulting cost separation precisely enough to be measured, and tests the approximation against exact posteriors. Its local bridge further admits a covariance analysis that delimits the scope of the variance-preserving correction.

\section{Methodology}
\label{sec:methodology}
\subsection{Problem Formulation}
\label{sec:problem}
Let $\X$ be  a compact spatial domain and $\U=C(\X)$ the space of continuous scalar fields on $\X$. We consider an unknown random field $u\in\U$ with prior distribution
\begin{equation}
    u\sim\mu,\qquad \mu\in\mathcal P(\U),
    \label{eq:data-distribution}
\end{equation}
where $\mathcal P(\U)$ denotes the set of probability measures on $\U$.

The field is observed at $m$ sensor locations $s_1,\ldots,s_m\in\X$ through the observation operator
\begin{equation}
    \A:\U\rightarrow\R^m,\qquad
    \A u=\bigl(u(s_1),\ldots,u(s_m)\bigr)^\top,
    \label{eq:observation-operator}
\end{equation}
with additive Gaussian noise modeled by
\begin{equation}
    \vy=\A u+\sigma\veps,\qquad
    \veps\sim\N(\vzero,I_m),
    \label{eq:obs-model}
\end{equation}
where $\sigma>0$ is the noise standard deviation and $I_m$ is the $m\times m$ identity matrix.

The observations constrain the prior through the likelihood. Under the Gaussian noise model, the negative log likelihood is given, up to an additive constant, by
\begin{equation}
    \Phi(u;\vy)=\frac{1}{2\sigma^2}\|\vy-\A u\|_2^2.
    \label{eq:field-likelihood}
\end{equation}
Bayes' rule defines the posterior distribution
$\pi^{\vy}\in\mathcal P(\U)$ as
\begin{equation}
    \pi^{\vy}(\mathrm du)
    \propto
    \exp[-\Phi(u;\vy)]\,\mu(\mathrm du),
    \label{eq:field-posterior}
\end{equation}
The objective is to generate samples from $\pi^{\vy}$ to characterize the spatial uncertainty remaining after conditioning on the observations.

Direct posterior sampling on a spatial grid requires a state whose dimension grows with the resolution, so the conditional updates become more expensive even when the number of observations is fixed. This motivates a latent representation that separates posterior sampling from field evaluation.

\subsection{Learned Latent Field Prior}
\label{sec:latent-representation}
\label{sec:latent-prior}

To separate posterior inference from spatial resolution, we represent each field realization $u\in\U$ by a latent variable $\vz\in\R^d$ through an encoder
\begin{equation}
    E_\phi:\U\rightarrow\R^d,
    \qquad
    \vz=E_\phi(u),
    \label{eq:encoder-map}
\end{equation}
where the latent dimension $d$ is independent of the spatial grid. The decoder $D_\psi:\R^d\rightarrow\U$ maps each latent variable to a continuous field, with value $D_\psi(\vz,x):=[D_\psi(\vz)](x)$ at $x\in\X$. We parameterize it using the Branch--Trunk representation of DeepONet \citep{lu2021deeponet}:
\begin{equation}
    D_\psi(\vz,x)
    =
    \mathbf b_\psi(\vz)^\top\mathbf h_\psi(x)
    +b_{\psi,0}(\vz),
    \label{eq:branch-trunk}
\end{equation}
where $\mathbf b_\psi(\vz),\mathbf h_\psi(x)\in\R^r$ denote the branch and trunk feature vectors, respectively, and $b_{\psi,0}(\vz)$ is a spatially constant offset. The branch features determine the field realization, whereas the trunk features characterize its spatial variation. For $n$ spatial locations, written as the ordered tuple $X_n=(x_1,\ldots,x_n)$ with $x_1,\ldots,x_n\in\X$, the decoder evaluates the field as
\begin{equation}
    D_\psi(\vz,X_n)=\bigl(D_\psi(\vz,x_1),\ldots,D_\psi(\vz,x_n)\bigr)^\top.
    \label{eq:decoder-evaluation}
\end{equation}
Thus, a single latent state defines a continuous field that can be evaluated at arbitrary spatial locations.
The encoder induces a latent distribution from the field prior $\mu$,
\begin{equation}
    \nu=(E_\phi)_{\#}\mu,
    \label{eq:encoded-latent-distribution}
\end{equation}
where $(E_\phi)_{\#}\mu$ denotes the pushforward measure of $\mu$ through $E_\phi$. To approximate $\nu$, we employ flow matching to learn a transport from a standard Gaussian reference distribution to the encoded latent distribution.

Let $\vx_0\sim\N(\vzero,I_d)$ and $\vx_1\sim\nu$ be independent samples. For $t\sim\mathcal U(0,1)$, we define the training path
\begin{equation}
    \vx_t
    =
    (1-t)\vx_0+t\vx_1+\sigma_{\min}\veps,
    \qquad
    \veps\sim\N(\vzero,I_d),
    \label{eq:latent-path}
\end{equation}
where $\veps$ is independent of $(\vx_0,\vx_1)$ and $\sigma_{\min}>0$ is a small noise level used for numerical regularization. The corresponding target velocity is given by $\vx_1-\vx_0$. Accordingly, a time-dependent velocity field $v_\theta:\R^d\times[0,1]\rightarrow\R^d$ is learned by minimizing
\begin{equation}
    \mathcal L_{\mathrm{FM}}
    =
    \E\left[
        \left\|
            v_\theta(\vx_t,t)-(\vx_1-\vx_0)
        \right\|_2^2
    \right].
    \label{eq:flow-matching-loss}
\end{equation}
The resulting velocity field defines the generative dynamics
\begin{equation}
    \frac{\mathrm d\vx(t)}{\mathrm dt}
    =
    v_\theta(\vx(t),t),
    \qquad
    \vx(0)\sim\N(\vzero,I_d).
    \label{eq:latent-flow-ode}
\end{equation}
Solving this equation from $t=0$ to $t=1$ yields the terminal latent state
\begin{equation}
    \vz=\vx(1)\sim\nu_\theta,
    \label{eq:learned-latent-prior}
\end{equation}
where $\nu_\theta$ approximates the encoded distribution $\nu$. We denote the covariance matrices of $\nu$ and $\nu_\theta$ by $\Sigma_z$ and $\Sigma_\theta$, respectively, noting that these matrices need not coincide.

Mapping samples from the learned latent prior through the decoder induces the field distribution
\begin{equation}
    \widehat\mu=(D_\psi)_{\#}\nu_\theta.
    \label{eq:learned-field-prior}
\end{equation}
Consequently, each latent realization generates a continuous field through the same decoder, thereby providing a common probabilistic representation across different spatial discretizations.

\subsection{Bayesian Conditioning in Latent Space}
\label{sec:latent-conditioning}
To incorporate the observations $\vy$ into the latent representation, we first express the likelihood \eqref{eq:field-likelihood} in terms of the latent variable $\vz$. Specifically, evaluating the decoded field at the sensor locations defines the latent observation map
\begin{equation}
    G_\psi(\vz)
    =
    \A D_\psi(\vz)
    =
    \bigl(D_\psi(\vz,s_1),\ldots,D_\psi(\vz,s_m)\bigr)^\top,
    \label{eq:latent-observation-map}
\end{equation}
where $G_\psi:\R^d\rightarrow\R^m$ maps each latent state to its predicted observations. The corresponding negative log likelihood in latent space is therefore
\begin{equation}
    \Phi_\psi(\vz;\vy)
    =
    \Phi(D_\psi(\vz);\vy)
    =
    \frac{1}{2\sigma^2}
    \left\|\vy-G_\psi(\vz)\right\|_2^2.
    \label{eq:latent-likelihood}
\end{equation}

Applying Bayes' rule to the learned latent prior $\nu_\theta$ defines the posterior distribution
\begin{equation}
    \pi_\theta^{\vy}(\mathrm d\vz)
    \propto
    \exp[-\Phi_\psi(\vz;\vy)]\,\nu_\theta(\mathrm d\vz),
    \label{eq:latent-posterior}
\end{equation}
Decoding samples from this posterior induces the corresponding field distribution
\begin{equation}
    \widehat\pi^{\vy}
    =
    (D_\psi)_{\#}\pi_\theta^{\vy},
    \label{eq:learned-field-posterior}
\end{equation}
Consequently, posterior inference over the field can be performed by sampling from $\pi_\theta^{\vy}$ in the finite dimensional latent space $\R^d$. Moreover, because the likelihood depends only on the field values at the $m$ sensor locations, the conditional updates are independent of the output grid.

\subsection{Latent Annealed Posterior Sampling}
\label{sec:laps}
\label{sec:annealed-sampling}

Direct sampling from $\pi_\theta^{\vy}$ is difficult when sparse observations concentrate its mass in complex regions of the learned prior. LAPS therefore uses an annealing procedure that gradually transforms a Gaussian reference distribution into this target.

Consider a terminal latent variable $\vz\sim\pi_\theta^{\vy}$ and the interpolation
\begin{equation}
    \vz_t=t\vz+(1-t)\veps,
    \qquad
    \veps\sim\N(\vzero,I_d),
    \qquad 0\le t\le1,
    \label{eq:posterior-path}
\end{equation}
where $\veps$ is independent of $\vz$. This interpolation defines the Gaussian kernel
\begin{equation}
    Q_t(\mathrm d\vz_t\mid\vz)=
    \begin{cases}
        \N\!\left(t\vz,(1-t)^2I_d\right)(\mathrm d\vz_t),
        & 0\le t<1,\\
        \delta_{\vz}(\mathrm d\vz_t),
        & t=1,
    \end{cases}
    \label{eq:posterior-bridge}
\end{equation}
where $\delta_{\vz}$ is the Dirac measure at $\vz$. Integrating this kernel against the posterior gives
\begin{equation}
    \pi_t^{\vy}(\mathrm d\vz_t)
    =
    \int_{\R^d}
    Q_t(\mathrm d\vz_t\mid\vz)\,
    \pi_\theta^{\vy}(\mathrm d\vz),
    \label{eq:annealed-marginal}
\end{equation}
which connects $\pi_0^{\vy}=\N(\vzero,I_d)$ to $\pi_1^{\vy}=\pi_\theta^{\vy}$.

The interpolation specifies the desired distributions, but does not provide a practical sampling procedure because its terminal variable is drawn from the posterior itself. An exact transition requires sampling this terminal variable conditioned on the current intermediate state. Bayes' rule gives
\begin{equation}
    \pi_\theta^{\vy}(\mathrm d\vz\mid\vz_t)
    \propto
    \exp[-\Phi_\psi(\vz;\vy)]
    Q_t(\vz_t\mid\vz)\,
    \nu_\theta(\mathrm d\vz),
    \label{eq:ideal-terminal-conditional}
\end{equation}
where $Q_t(\vz_t\mid\vz)$ denotes the Gaussian density associated with \eqref{eq:posterior-bridge} for $t<1$. An exact draw from this conditional distribution, followed by re-bridging to a later time, produces the corresponding intermediate marginal. This conditional involves the implicitly represented latent prior $\nu_\theta$, so direct sampling is generally intractable. LAPS approximates this conditional update using the learned prior flow and a local Gaussian distribution.

We discretize the annealing interval as
\begin{equation}
    0=t_0<t_1<\cdots<t_K=1,
    \label{eq:annealing-schedule}
\end{equation}
where $K$ is the number of annealing levels. Let $\vz_k$ denote the numerical state at level $k$, initialized from
\begin{equation}
    \vz_0\sim\N(\vzero,I_d).
    \label{eq:laps-initial-state}
\end{equation}
The implemented noise width is defined by
\begin{equation}
    \beta_k=\sqrt{(1-t_k)^2+\sigma_{\min}^2},
    \qquad k=0,\ldots,K-1,
    \label{eq:bridge-scale}
\end{equation}
with $\sigma_{\min}$ the residual noise level of \eqref{eq:latent-path}. The recursion has $K$ levels, $k=0,\ldots,K-1$; the terminal time $t_K=1$ is the target of every transport and is not itself a level.

To approximate the terminal conditional update, we use the learned prior dynamics to obtain an endpoint from the current state. Let $T^\theta_{a,b}:\R^d\rightarrow\R^d$ denote the flow map generated by \eqref{eq:latent-flow-ode} from time $a$ to $b$. The transported endpoint is
\begin{equation}
    \bar{\vz}_k=T^\theta_{t_k,1}(\vz_k),
    \label{eq:endpoint-transport}
\end{equation}
which serves as the anchor of the local update and is computed with $S$ explicit Euler steps. The transport follows the learned prior dynamics, not the posterior interpolation, so the endpoint is approximate but requires no draw from the terminal conditional distribution.

The anchor alone does not carry the uncertainty of the conditional update, so a local Gaussian distribution is placed around it. Its construction must account for the variability already carried by the transported states: an independent Gaussian perturbation added with unit anchor weight inflates the covariance when the anchor already matches the learned prior covariance. This motivates a variance-preserving (VP) local prior of the form
\begin{equation}
    q_k(\mathrm d\vz\mid\bar{\vz}_k)
    =
    \N\!\left(
        \mathrm d\vz;
        \alpha_k\bar{\vz}_k,
        \lambda_k^2\Sigma_z
    \right),
    \label{eq:vp-local-prior}
\end{equation}
where $\alpha_k$ and $\lambda_k$ are the anchor and noise coefficients, respectively, and $\Sigma_z$ is the encoded latent covariance. The anchor is scaled toward the origin, which is the mean of the standardized codes.

The coefficients are chosen to allocate a single covariance budget between the transported anchor and the Gaussian perturbation:
\begin{equation}
    \alpha_k^2+\lambda_k^2=1.
    \label{eq:vp-condition}
\end{equation}
Under matched anchor and kernel covariances, this budget preserves the covariance of the learned prior before likelihood correction. To determine the individual coefficients, we retain the endpoint-to-noise ratio of the implemented annealing path, $\alpha_k/\lambda_k=t_k/\beta_k$. The resulting coefficients are
\begin{equation}
    \alpha_k=\frac{t_k}{\sqrt{t_k^2+\beta_k^2}},
    \qquad
    \lambda_k=\frac{\beta_k}{\sqrt{t_k^2+\beta_k^2}},
    \label{eq:bridge-coeffs}
\end{equation}
for $k=0,\ldots,K-1$. For a general prior, this ratio is a design choice rather than a consequence of Bayesian conditioning. The original kernel of \citet{shi2026faps} uses unit anchor weight and noise width $1-t_k$. In the implementation, the VP width is bounded below by $\lambda_{\min}>0$ for numerical stability, so \eqref{eq:vp-condition} holds exactly only at levels where this floor is inactive.

With the local prior specified, the observation likelihood can be incorporated into the terminal update. The resulting local posterior is
\begin{equation}
    \widetilde{\pi}_k(\mathrm d\vz\mid\bar{\vz}_k,\vy)
    \propto
    \exp[-\Phi_\psi(\vz;\vy)]
    q_k(\mathrm d\vz\mid\bar{\vz}_k).
    \label{eq:local-target}
\end{equation}
We sample from this distribution using the preconditioned Metropolis-adjusted Langevin algorithm (MALA) \citep{roberts1996exponential}. Given a current state $\vz$, a candidate is generated by
\begin{equation}
    \begin{aligned}
    \vz'={}&\vz-\frac{\tau_k}{2}P_k
    \bigl[
        \nabla_{\vz}\Phi_\psi(\vz;\vy)
        +\lambda_k^{-2}\Sigma_z^{-1}
        (\vz-\alpha_k\bar{\vz}_k)
    \bigr]\\
    &+\sqrt{\tau_k}P_k^{1/2}\veps,
    \qquad \veps\sim\N(\vzero,I_d),
    \end{aligned}
    \label{eq:mala-proposal}
\end{equation}
where $\tau_k>0$ is the step size and $P_k\succ0$ is a positive-definite preconditioner that is fixed within level $k$; we use a Gauss--Newton form and tune $\tau_k$ on held-out scenarios (\Cref{app:mala}). The drift contains the gradients of the observation likelihood and the VP local prior. A Metropolis--Hastings correction accepts or rejects the candidate according to the local posterior. Starting from a draw of $q_k$, we perform $L$ MALA steps to obtain the observation-corrected endpoint $\vz_k^+$.

The corrected endpoint is then returned to the annealing path at the next level through
\begin{equation}
    \vz_{k+1}
    =
    t_{k+1}\vz_k^+
    +
    \beta_{k+1}\veps_k,
    \qquad
    \veps_k\sim\N(\vzero,I_d),
    \label{eq:rebridge}
\end{equation}
where $\veps_k$ is an independent standard Gaussian vector and $k=0,\ldots,K-2$. The last level, $k=K-1$, is not re-bridged: its corrected endpoint is retained as the terminal sample, $\vz_K=\vz_{K-1}^+$, so no noise is added at $t_K=1$. The resulting latent samples are decoded once at the requested spatial locations.

LAPS thus replaces the exact terminal conditional draw with prior transport, a VP local prior and finite MALA updates, and its transitions approximate the ideal annealing construction. \Cref{alg:laps} summarizes the procedure.

\begin{algorithm}[t]
\caption{Latent Annealed Posterior Sampling}
\label{alg:laps}
\begin{algorithmic}[1]
\Require Observations $\vy$ at the sensors $s_1,\ldots,s_m$; output locations $X_n$;
decoder $D_\psi$; flow map $T^\theta$, computed with $S$ Euler steps;
covariance $\Sigma_z$; schedule $\{t_k\}_{k=0}^{K}$; step sizes $\{\tau_k\}$ and
preconditioners $\{P_k\}$; number of particles $N$; MALA steps $L$.
\Ensure Posterior field samples $\{u^{(i)}\}_{i=1}^{N}$.

\State Compute $\{(\alpha_k,\lambda_k)\}_{k=0}^{K-1}$
from \eqref{eq:bridge-coeffs}, with $\lambda_k$ floored at $\lambda_{\min}$
(\Cref{app:schedule}).

\For{$i=1,\ldots,N$}
    \State Draw $\vz_0\sim\N(\vzero,I_d)$.
    \For{$k=0,\ldots,K-1$}
        \State
        $\bar{\vz}_k
        \gets
        T_{t_k,1}^{\theta}\!\left(\vz_k\right)$.

        \State Apply $L$ MALA steps \citep{roberts1996exponential}
        targeting \eqref{eq:local-target}, started from \eqref{eq:chain-start},
        to obtain $\vz_k^+$.

        \If{$k<K-1$}
            \State Draw $\veps_k\sim\N(\vzero,I_d)$.
            \State
            $\vz_{k+1}
            \gets
            t_{k+1}\vz_k^+
            +
            \beta_{k+1}\veps_k$.
        \Else
            \State
            $\vz_K\gets\vz_k^+$.
        \EndIf
    \EndFor

    \State
    $u^{(i)}
    \gets
    D_\psi\!\left(\vz_K,X_n\right)$.
\EndFor

\State \Return $\{u^{(i)}\}_{i=1}^{N}$.
\end{algorithmic}
\end{algorithm}
\section{Theory and Analysis}
\label{sec:theory}

This section states what the construction guarantees and where the guarantees stop: latent conditioning induces a coherent random-field posterior, ideal annealing samples it exactly, the VP bridge controls the covariance of a single local update, and the evaluation cost separates. These statements fix the target and the scope of the method; none is an accuracy guarantee for the implemented sampler.

\subsection{Coherent Latent Field Representation}
\label{sec:analysis-representation}

For the decoded representation to define a random-field model, its finite-dimensional distributions must be mutually consistent, and conditioning in latent space must induce the posterior under the learned field prior.

\begin{proposition}[Coherent field prior and posterior]
\label[proposition]{thm:process-consistency}
Let $\X$ be a compact metric space and $\U=C(\X)$ carry the supremum norm. Assume that $\nu_\theta$ is a Borel probability measure on $\R^d$ and that $D_\psi:\R^d\to\U$ is Borel measurable. Write $\mathcal Z_\theta(\vy)=\int_{\R^d}\exp[-\Phi_\psi(\vz;\vy)]\,\nu_\theta(\mathrm d\vz)$ for the normalizing constant. Then $0<\mathcal Z_\theta(\vy)<\infty$, and the learned field prior $\widehat\mu$ and posterior $\widehat\pi^{\vy}$ are Borel probability measures on $\U$. For any $n$ spatial locations $X_n=(x_1,\ldots,x_n)$, their finite-dimensional distributions
\begin{equation}
    \widehat\mu_{X_n}
    =
    \bigl(D_\psi(\cdot,X_n)\bigr)_{\#}\nu_\theta,
    \qquad
    \widehat\pi_{X_n}^{\vy}
    =
    \bigl(D_\psi(\cdot,X_n)\bigr)_{\#}\pi_\theta^{\vy}
\end{equation}
are consistent under permutations and restrictions to subsets of locations, and uniquely determine the corresponding field measures. Moreover,
\begin{equation}
    \frac{\mathrm d\widehat\pi^{\vy}}{\mathrm d\widehat\mu}(u)
    =
    \frac{\exp[-\Phi(u;\vy)]}{\mathcal Z_\theta(\vy)}.
    \label{eq:decoded-field-posterior}
\end{equation}
\end{proposition}

Conditioning the latent prior and decoding its posterior samples is thus equivalent to Bayesian conditioning under the learned field prior, without requiring the decoder to be injective. The proof is in \Cref{app:process}.

\begin{remark}[Scope of the learned posterior]
The proposition establishes consistency of the learned field model, not its approximation accuracy. Exact sampling from $\pi_\theta^{\vy}$ produces $\widehat\pi^{\vy}$, which coincides with the original posterior $\pi^{\vy}$ when $\widehat\mu=\mu$ and may differ from it otherwise. Finite-step LAPS sampling introduces an additional approximation to this learned-model target.
\end{remark}

\subsection{Exact Posterior Sampling under Ideal Annealing}
\label{sec:analysis-annealing}

The intermediate distributions \eqref{eq:annealed-marginal} connect the Gaussian reference to the latent posterior $\pi_\theta^{\vy}$. An ideal transition draws a terminal state conditioned on the current annealed state and then applies the bridge kernel at a later time.

\begin{proposition}[Ideal annealed transition]
\label[proposition]{prop:ideal-transition}
Let $0\le t'<t\le1$ and $\vz_{t'}\sim\pi_{t'}^{\vy}$. Draw $\vz$ from the exact terminal conditional distribution $\pi_\theta^{\vy}(\cdot\mid\vz_{t'})$, followed by $\vz_t\sim Q_t(\cdot\mid\vz)$. Then
\begin{equation}
    \vz_t\sim\pi_t^{\vy}.
\end{equation}
\end{proposition}

This is the ideal annealing principle of \citet[Proposition~1]{shi2026faps}, stated for the latent posterior (\Cref{app:ideal-transition}). Repeating the transition along the schedule yields $\vz_K\sim\pi_\theta^{\vy}$, and decoding gives $\widehat\pi^{\vy}$ by \Cref{thm:process-consistency}. Neither an explicit prior density nor a Gaussian prior is required.

\begin{remark}[Ideal and practical transitions]
\label[remark]{rem:ideal-practical}
The proposition relies on exact terminal conditional sampling. LAPS instead uses learned prior transport, a local Gaussian approximation and finite MALA updates, with further deviations from numerical flow integration, residual noise and the width floor, so the proposition does not guarantee exactness of the implemented sampler. A special correspondence arises when the learned latent prior is $\N(\vzero,I_d)$, $\Sigma_z=I_d$, the independent-coupling flow transport is exact, and both residual noise and the width floor are absent. Under these conditions, the VP local prior $q_k(\cdot\mid\bar{\vz}_k)$ coincides with the conditional prior of $\vz$ given $\vz_{t_k}=\vz_k$, which is proportional to $Q_{t_k}(\vz_k\mid\vz)\,\nu_\theta(\mathrm d\vz)$, so exact local sampling yields the target posterior (\Cref{app:ideal-transition}). This correspondence does not generally extend to arbitrary Gaussian covariances or non-Gaussian latent priors.
\end{remark}

\subsection{Local Covariance Preservation and Mismatch}
\label{sec:analysis-bridge}

The transported endpoint already carries the uncertainty of the learned prior, so an independent Gaussian perturbation around it can increase the covariance. The original kernel of \citet{shi2026faps} uses unit anchor weight and noise width $1-t_k$: if the anchor and kernel covariances both equal the covariance $\Sigma_\theta$ of the learned latent prior, the local draw has covariance $[1+(1-t_k)^2]\Sigma_\theta$ before any likelihood correction.

The VP coefficients \eqref{eq:bridge-coeffs} instead satisfy the budget \eqref{eq:vp-condition}. Consider a draw from the local prior before likelihood correction:
\begin{equation}
    \mathbf w_k
    =
    \alpha_k\bar{\vz}_k
    +
    \lambda_k\Sigma_z^{1/2}\veps_k,
    \qquad
    \veps_k\sim\N(\vzero,I_d),
\end{equation}
where $\veps_k$ is independent of $\bar{\vz}_k$, and define the covariance mismatch of the transported endpoint by
\begin{equation}
    \Delta_k
    =
    \operatorname{Cov}(\bar{\vz}_k)-\Sigma_\theta.
\end{equation}

\begin{proposition}[Local covariance error]
\label[proposition]{thm:covariance-preservation}
Assume that the learned prior and transported endpoint have finite second moments, the Gaussian perturbation is independent of the endpoint, and the unfloored coefficients satisfy $\alpha_k^2+\lambda_k^2=1$. For any positive semidefinite covariance matrix $\Sigma_z$,
\begin{equation}
    \operatorname{Cov}(\mathbf w_k)-\Sigma_\theta
    =
    \alpha_k^2\Delta_k
    +
    \lambda_k^2(\Sigma_z-\Sigma_\theta).
    \label{eq:bridge-covariance-mismatch}
\end{equation}
In particular, if $\Delta_k=0$ and $\Sigma_z=\Sigma_\theta$, then
\begin{equation}
    \operatorname{Cov}(\mathbf w_k)=\Sigma_\theta.
    \label{eq:covariance-preservation}
\end{equation}
No Gaussian assumption on the learned prior or transported endpoint is required.
\end{proposition}

The identity is a second-moment calculation (\Cref{app:covariance-preservation}). Its two terms are the covariance mismatch of the transported endpoint and the difference between the encoded covariance $\Sigma_z$ used in the kernel and the covariance $\Sigma_\theta$ of the learned prior. When both vanish, the VP coefficients remove the inflation of the unit-weight construction, whether or not the latent prior is Gaussian.

\begin{remark}[Scope of covariance preservation]
\label[remark]{rem:scope-covariance}
The proposition concerns a local prior draw before likelihood correction, with covariance taken over both the random endpoint and the independent Gaussian perturbation. It does not establish preservation of the posterior covariance or invariance of the complete annealing recursion. The corresponding mean satisfies $\E[\mathbf w_k]=\alpha_k\E[\bar{\vz}_k]$, so matching covariance alone does not guarantee preservation of the first moment. An active width floor also modifies the covariance budget, contributing the additional term $(\alpha_k^2+\lambda_k^2-1)\Sigma_\theta$ to the covariance error. For non-Gaussian distributions, the perturbation can change higher-order structure even when covariance is preserved. In particular, whenever the fourth-order cumulants exist, they scale by $\alpha_k^4$ relative to those of the transported endpoint. Subsequent nonlinear transport may introduce further covariance mismatch.
\end{remark}

\begin{remark}[Gaussian priors]
\label[remark]{rem:gaussian-closed}
One case is closed. Call the transport exact if $T^\theta_{t,1}$ maps the law of $t\vz+(1-t)\veps$, with $\vz\sim\nu_\theta$ and $\veps\sim\N(\vzero,I_d)$ independent, to $\nu_\theta$, and suppose that the learned prior is a centered Gaussian, $\Sigma_z=\Sigma_\theta$, the floor is inactive and $\sigma_{\min}=0$. If $\bar{\vz}_k\sim\nu_\theta$, then $\mathbf w_k\sim\nu_\theta$ and the re-bridged state has the law that exact transport returns to $\nu_\theta$. Since $\bar{\vz}_0\sim\nu_\theta$ by definition, every level of the likelihood-free recursion returns the prior. For a non-Gaussian prior only the covariance of $\mathbf w_k$ is matched, and this induction fails at its first step.
\end{remark}

\subsection{Computational Cost}
\label{sec:analysis-cost}

With the latent dimension $d$, the number of sensors $m$, the schedule $(K,L,S)$, the ensemble size $N$ and the network architectures fixed, the evaluation time for $n$ output locations separates as
\begin{equation}
    T_{\mathrm{eval}}(n)
    =
    T_{\mathrm{cond}}(d,m,K,L,S)
    +T_{\mathrm{dec}}(n)
    +T_{\mathrm{score}}(n),
    \label{eq:cost-split}
\end{equation}
where $T_{\mathrm{cond}}$ is the cost of the annealed recursion, $T_{\mathrm{dec}}$ that of the final decoding and $T_{\mathrm{score}}$ that of computing metrics. The recursion makes $NKS$ velocity evaluations in $\R^d$ and $NKL$ MALA updates, each of which evaluates the decoder and its gradient at the $m$ sensors only, so $T_{\mathrm{cond}}$ does not involve $n$. Decoding evaluates the trunk once at the $n$ output locations and contracts the $N$ branch vectors with its rank-$r$ features, at a cost of $O(Nnr)$ beyond the trunk evaluation. Refining the output grid therefore changes $T_{\mathrm{dec}}$ and $T_{\mathrm{score}}$ and leaves $T_{\mathrm{cond}}$ unchanged. This describes how the cost is composed. It asserts neither an end-to-end time independent of resolution nor a lower bound for field-space samplers; in implementations that keep the Markov state and its covariance on the output grid, every repeated update involves $n$.
\section{Experiments}
\label{sec:experiments}
The experiments proceed from exact references to a field without one. Gaussian benchmarks test accuracy against analytic posteriors (\Cref{sec:exp-matern}), a mixture benchmark tests a non-Gaussian analytic posterior and the retention of two plausible configurations (\Cref{sec:exp-bimodal}), and Navier--Stokes vorticity tests predictive accuracy and marginal calibration where no reference posterior exists (\Cref{sec:ns}). A resolution sweep then examines the computational separation (\Cref{sec:resolution}), and likelihood-free diagnostics and ablations examine the bridge mechanism and the annealing schedule (\Cref{sec:exp-ablation}).

\subsection{Experimental Setup}
\label{sec:experimental-design}

Three benchmarks are one-dimensional and have an exact reference posterior: the stationary Mat\'ern-\(3/2\) and non-stationary Gibbs Gaussian-process benchmarks of \citet{shi2026faps}, and a bimodal mixture (\Cref{sec:exp-bimodal}). The fourth infers Navier--Stokes vorticity on a \(64\times64\) torus from \(64\) point observations. Every method draws \(128\) posterior samples for each of \(100\) test scenarios, and the exact posterior is sampled at the same size, providing a finite-ensemble reference for each benchmark and metric. On the one-dimensional benchmarks LAPS runs at the published FAPS budget (\(K=40\), \(S=20\), \(L=50\)), so both samplers spend \(2000\) local updates per scenario. On Navier--Stokes LAPS uses \(K=80\) and \(L=100\); its \(8000\) updates act on a \(128\)-dimensional latent state and the \(2000\) updates of FAPS on the \(4096\)-point field, so the update counts are not matched on this benchmark; a control at the FAPS count is reported in \Cref{app:matched-budget}. Step sizes are tuned on held-out scenarios and frozen. We report sliced Wasserstein distance (SWD), maximum mean discrepancy (MMD) and the errors of the posterior mean and pointwise standard deviation where an exact posterior exists, and CRPS, spread--skill ratio (SSR), relative \(L^2\) error and calibration error (CE) on Navier--Stokes. Trained methods are repeated over five training seeds. Settings and metric definitions are in \Cref{app:parameters,app:metrics}.

The baselines are amortized neural processes---CNP, ANP and FlowNP
\citep{garnelo2018cnp,kim2019anp,abuhamad2025flownp}, trained on the same fields---which
bound what amortization alone achieves, and FAPS \citep{shi2026faps}, the
field-space sampler from which the annealing construction is taken, run with
its released weights where they exist and always with its original sampler
at its published settings (\Cref{app:reproduction}). FAPS is the method closest in construction to LAPS, and the Gaussian benchmarks are those of its evaluation; other function-space samplers are outside the scope of this comparison. This comparison is between complete methods and does not isolate the contribution of the latent representation; the bridge coefficients are compared separately within LAPS (\Cref{sec:exp-ablation}).

\subsection{Gaussian Posterior Benchmarks}
\label{sec:exp-matern}

We first assess posterior distribution accuracy on stationary Mat\'ern and non-stationary Gibbs Gaussian process benchmarks on \(\X=[0,1]\). Their Gaussian priors and linear Gaussian observation model give analytic conditional posteriors, allowing direct comparison with exact references. Each scenario contains seven point observations with noise standard deviation \(0.1\), and the comparison uses \(n=128\) evaluation points; the covariance models and settings are in \Cref{app:benchmark-construction,app:parameters}.

In \Cref{tab:matern}, the LAPS SWD on the stationary Mat\'ern benchmark is \(0.0614\), close to the \(0.0603\) obtained from an independent ensemble of exact posterior draws, while FAPS gives \(0.1397\). The non-stationary Gibbs benchmark and the other three metrics show the same ordering. At this ensemble size an SWD above zero reflects the finite-ensemble reference, not a fixed bias; closeness to that reference does not in turn establish equality with the exact posterior. Recorded evaluation times are reported for completeness; the device mismatch in \Cref{app:hardware} prevents interpreting them as a controlled speed comparison.

\begin{table}[t]
  \centering
  \small
  \setlength{\tabcolsep}{3.5pt}
  \caption{Posterior accuracy on the stationary Mat\'ern and non-stationary Gibbs benchmarks at \(n=128\). The reference row uses an independent ensemble of exact posterior draws. \(\dagger\) Times include scoring and come from different devices; they are recorded evaluation times, not a controlled speed comparison (\Cref{app:hardware}).}
  \label{tab:matern}
\small
\begin{tabular}{lccccr}
\toprule
Method & SWD $\downarrow$ & MMD $\downarrow$ & $\mathrm{RMSE}_m$ $\downarrow$ & $\mathrm{MAE}_s$ $\downarrow$ & s/case$^{\dagger}$ \\
\midrule
\multicolumn{6}{l}{\emph{Mat\'ern (stationary)}} \\
\addlinespace[0.15em]
CNP & $0.2087_{\pm 0.0072}$ & $0.2002_{\pm 0.0098}$ & $0.1521_{\pm 0.0084}$ & $0.0671_{\pm 0.0022}$ & 0.07 \\
ANP & $0.1380_{\pm 0.0019}$ & $0.1200_{\pm 0.0035}$ & $0.0895_{\pm 0.0032}$ & $0.0359_{\pm 0.0007}$ & 0.08 \\
FlowNP & $0.2149_{\pm 0.0027}$ & $0.1978_{\pm 0.0051}$ & $0.1367_{\pm 0.0047}$ & $0.1403_{\pm 0.0016}$ & 1.21 \\
\cmidrule(lr){1-6}
FAPS & $0.1397_{\pm 0.0009}$ & $0.1321_{\pm 0.0010}$ & $0.1206_{\pm 0.0008}$ & $0.0269_{\pm 0.0006}$ & 12.52 \\
LAPS (ours) & $\mathbf{0.0614}_{\pm 0.0007}$ & $\mathbf{0.0147}_{\pm 0.0012}$ & $\mathbf{0.0266}_{\pm 0.0011}$ & $\mathbf{0.0133}_{\pm 0.0005}$ & 5.49 \\
Reference (exact draws) & $0.0603_{\pm 0.0011}$ & $0.0108_{\pm 0.0023}$ & $0.0239_{\pm 0.0014}$ & $0.0127_{\pm 0.0006}$ & -- \\
\addlinespace[0.5em]\midrule
\multicolumn{6}{l}{\emph{Gibbs (non-stationary)}} \\
\addlinespace[0.15em]
CNP & $0.3034_{\pm 0.0076}$ & $0.2904_{\pm 0.0090}$ & $0.2352_{\pm 0.0080}$ & $0.1037_{\pm 0.0058}$ & 0.08 \\
ANP & $0.1902_{\pm 0.0032}$ & $0.1628_{\pm 0.0047}$ & $0.1244_{\pm 0.0050}$ & $0.0449_{\pm 0.0002}$ & 0.08 \\
FlowNP & $0.2719_{\pm 0.0133}$ & $0.2520_{\pm 0.0177}$ & $0.1794_{\pm 0.0162}$ & $0.1628_{\pm 0.0038}$ & 1.28 \\
\cmidrule(lr){1-6}
FAPS & $0.1695_{\pm 0.0014}$ & $0.1398_{\pm 0.0009}$ & $0.1366_{\pm 0.0011}$ & $0.0413_{\pm 0.0009}$ & 12.03 \\
LAPS (ours) & $\mathbf{0.0762}_{\pm 0.0006}$ & $\mathbf{0.0180}_{\pm 0.0027}$ & $\mathbf{0.0317}_{\pm 0.0015}$ & $\mathbf{0.0147}_{\pm 0.0004}$ & 5.86 \\
Reference (exact draws) & $0.0744_{\pm 0.0010}$ & $0.0149_{\pm 0.0019}$ & $0.0294_{\pm 0.0012}$ & $0.0140_{\pm 0.0005}$ & -- \\
\bottomrule
\end{tabular}

\end{table}
\FloatBarrier

\begin{figure}[t]
  \centering
  \includegraphics[width=\linewidth]{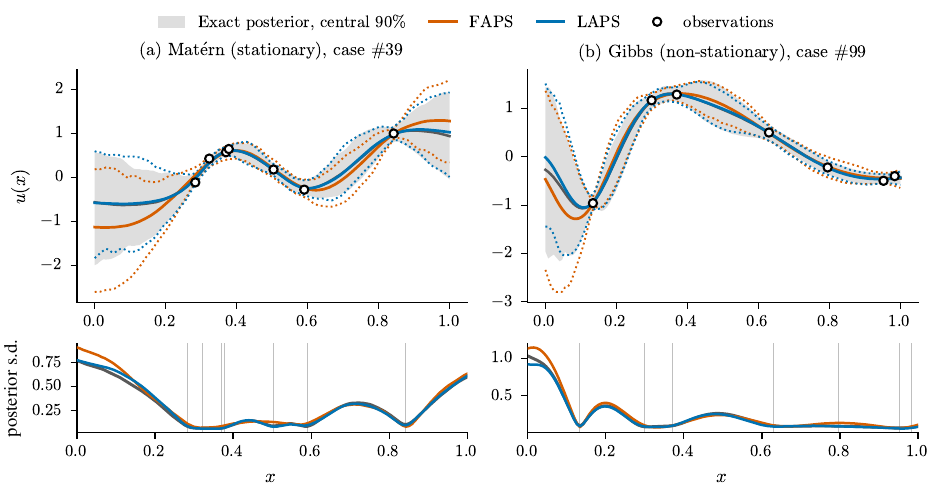}
  \caption{Representative posterior estimates for the Mat\'ern and Gibbs benchmarks at \(n=128\). The gray band denotes the central \(90\%\) interval of the analytic posterior; FAPS and LAPS intervals are empirical posterior quantiles. The lower panels show the pointwise posterior standard deviation.}
  \label{fig:matern_case}
\end{figure}

\Cref{fig:matern_case} presents a representative posterior for each benchmark, corresponding to the test scenario with the median LAPS SWD. In both cases, the LAPS interval closely tracks the analytic posterior across the domain, while its pointwise standard deviation agrees well with the reference profile. FAPS captures the overall posterior structure but exhibits larger deviations from the analytic reference.

At \(n=512\), where the observation locations are mapped to the finer grid, the LAPS SWD stays within \(2\%\) of the exact-draw reference on both benchmarks (\Cref{app:benchmark-construction}).
\FloatBarrier

\subsection{Non-Gaussian Mixture Posterior}
\label{sec:exp-bimodal}

\let\lapsSavedBottomFraction\bottomfraction
\renewcommand{\bottomfraction}{0.45}

The preceding Gaussian benchmarks have unimodal posteriors and therefore do not test mode recovery. We next consider a field on \(\X=[0,1]\) with two possible localized bump locations. Five point observations with noise standard deviation \(0.05\) leave both structures plausible. The conditional posterior is an exact two-component Gaussian mixture with closed-form weights; across the \(100\) test scenarios, the weight of the first component ranges from \(0.10\) to \(0.89\), covering both nearly balanced and strongly asymmetric cases. The field construction is given in \Cref{app:benchmark-construction}.

The comparison uses \(n=128\) and the sampler configuration of the Gaussian benchmarks; the latent flow is wider and trained for longer than on the Gaussian benchmarks, a choice fixed from likelihood-free diagnostics of the learned prior (\Cref{app:settings}). Three template-based metrics complement the distances to the exact posterior. The valley mass is the fraction of draws that lie between the two configurations, measured through the fitted amplitudes of the two bumps; it separates two distinct modes from one broad ensemble. ModeCov is the fraction of scenarios in which both template regions receive at least \(5\%\) of the samples, and ModeW-MAE compares the mass assigned to the first region with the exact mixture weight. The reference row of \Cref{tab:bimodal} scores independent ensembles of exact posterior draws and gives the finite-ensemble floor of each column (\Cref{app:metrics}).

\begin{figure}[tbp]
  \centering
  \includegraphics[width=\linewidth]{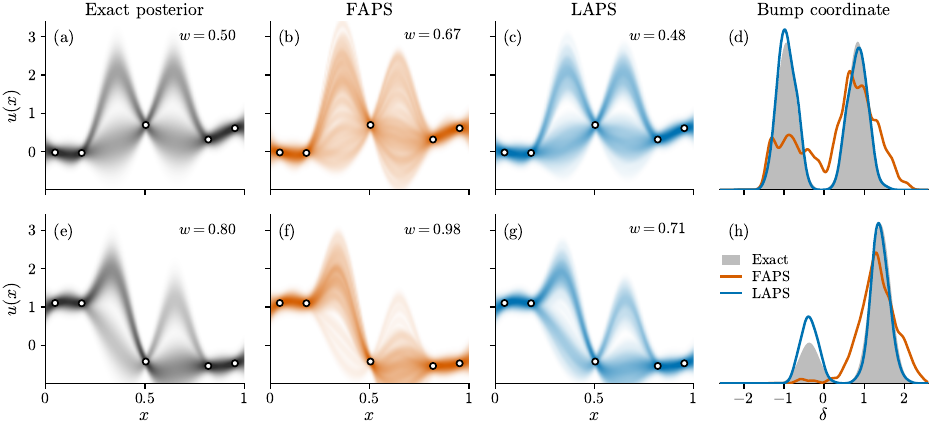}
  \caption{Posterior on the bimodal benchmark. Rows: two test scenarios, in which the exact weight of the configuration with the bump at \(x=0.35\) is \(0.50\) (top) and \(0.80\) (bottom). (a--c, e--g) Pointwise posterior density of \(u(x)\) for the exact posterior, FAPS and LAPS; darker is denser, on a common square-root scale within each row. Circles are the observations and \(w\) is the mass assigned to the first configuration. (d, h) Density of the bump coordinate \(\delta\), the fitted amplitude of the bump at \(0.35\) minus that of the bump at \(0.65\). Sampler densities pool the draws of the five training seeds, and all densities are smoothed with the same Gaussian kernel.}
  \label{fig:bimodal-modes}
\end{figure}

LAPS has the smallest error in every column of \Cref{tab:bimodal}. Its draws separate into the two configurations as exact draws do, with a valley mass of \(0.045\) against \(0.042\), and both configurations are represented in every scenario, whereas FAPS loses one of them in \(27\%\) of the scenarios. The distances stay above the exact-draw floor, by \(37\%\) in SWD. One identifiable source is that the sampled mode weights are flattened toward one half: in every seed the logit of the sampled weight grows about \(0.65\) times as fast as that of the exact weight, with an offset that varies between seeds (\Cref{app:mode-weights}). FAPS errs in the opposite direction and sharpens the weights. The template metrics do not identify generating-component labels for individual draws.
\begin{table}[tbp]
  \centering
  \caption{Posterior accuracy and mode structure on the bimodal benchmark at \(n=128\): five training seeds, mean \(\pm\) standard deviation. Distributional errors use exact posterior references. ModeW-MAE and ModeCov use a fixed template assignment; the valley mass is the fraction of draws between the two configurations. The reference row scores five independent ensembles of exact posterior draws (\Cref{app:metrics}).}
  \label{tab:bimodal}
  \resizebox{\linewidth}{!}{
\small
\begin{tabular}{lcccccc}
\toprule
 & \multicolumn{3}{c}{distance to the exact posterior} & \multicolumn{3}{c}{template regions} \\
\cmidrule(lr){2-4}\cmidrule(lr){5-7}
Method & SWD $\downarrow$ & MMD $\downarrow$ & $\mathrm{RMSE}_m$ $\downarrow$ & ModeW-MAE $\downarrow$ & ModeCov $\uparrow$ & Valley $\to0.04$ \\
\midrule
CNP & $0.4666_{\pm 0.0292}$ & $0.4124_{\pm 0.0223}$ & $0.3137_{\pm 0.0306}$ & $0.3636_{\pm 0.0666}$ & $0.378_{\pm 0.199}$ & $0.703_{\pm 0.172}$ \\
ANP & $0.3579_{\pm 0.0044}$ & $0.3255_{\pm 0.0072}$ & $0.1721_{\pm 0.0069}$ & $0.2860_{\pm 0.0093}$ & $0.320_{\pm 0.029}$ & $0.685_{\pm 0.038}$ \\
FlowNP & $0.3267_{\pm 0.1306}$ & $0.2476_{\pm 0.1124}$ & $0.1934_{\pm 0.1117}$ & $0.2103_{\pm 0.1348}$ & $0.856_{\pm 0.130}$ & $0.092_{\pm 0.034}$ \\
\cmidrule(lr){1-7}
FAPS & $0.2693_{\pm 0.0007}$ & $0.1750_{\pm 0.0006}$ & $0.1964_{\pm 0.0006}$ & $0.1502_{\pm 0.0013}$ & $0.730_{\pm 0.007}$ & $0.099_{\pm 0.003}$ \\
LAPS (ours) & $\mathbf{0.1499}_{\pm 0.0090}$ & $\mathbf{0.0529}_{\pm 0.0079}$ & $\mathbf{0.0678}_{\pm 0.0056}$ & $\mathbf{0.0661}_{\pm 0.0071}$ & $\mathbf{1.000}_{\pm 0.000}$ & $\mathbf{0.045}_{\pm 0.001}$ \\
\midrule
Reference (exact draws) & $0.1096_{\pm 0.0034}$ & $0.0222_{\pm 0.0031}$ & $0.0327_{\pm 0.0024}$ & $0.0316_{\pm 0.0018}$ & $1.000_{\pm 0.000}$ & $0.042_{\pm 0.001}$ \\
\bottomrule
\end{tabular}
}
\end{table}
\FloatBarrier
\Cref{fig:bimodal-modes} shows the posterior itself for two scenarios, one with balanced exact weights and one with weights of \(0.80\) and \(0.20\). In function space the exact posterior has two branches at each bump location. The LAPS density reproduces both, whereas the FAPS density is diffuse and, in the asymmetric scenario, has almost lost the minority branch. Along the bump coordinate the LAPS modes coincide with the exact ones in position and width; in the asymmetric scenario the minority mode receives \(0.29\) of the mass instead of \(0.20\), the flattening noted above.
\FloatBarrier

\subsection{Two-Dimensional Navier--Stokes}
\label{sec:ns}

\let\bottomfraction\lapsSavedBottomFraction

We next assess predictive accuracy and pointwise uncertainty calibration on a non-Gaussian two-dimensional field benchmark: incompressible Navier--Stokes vorticity snapshots on a periodic domain. Each scenario contains \(64\) noisy point observations of a \(64\times64\) field. The data source and benchmark construction are given in \Cref{app:ns-data}.

The conditional posterior is not available analytically, so we evaluate ensembles against the realized test fields. CRPS and posterior-mean relative \(L^2\) error assess predictive and reconstruction accuracy, CE assesses marginal calibration, and SSR compares spread with reconstruction error. LAPS uses \(d=128\), with the posterior-sample and annealing budgets specified in \Cref{sec:experimental-design} and the full settings in \Cref{app:parameters}.

\Cref{tab:e4ns} shows an accuracy--calibration trade-off. FAPS attains lower CRPS and relative \(L^2\) error, by 23\% and 15\%. LAPS gives better aggregate marginal calibration: its SSR is \(0.989\), compared with \(1.136\) for FAPS, and its CE is \(0.0047\), compared with \(0.0352\). The comparison does not rest on the larger number of LAPS updates: restricted to the \(2000\) updates of FAPS, LAPS has a CE of \(0.012\) and an SSR of \(1.02\), with CRPS and relative \(L^2\) error within \(3\%\) of the reported values (\Cref{app:matched-budget}). Nor is the accuracy difference explained by reconstruction error: the autoencoder reconstructs held-out fields to a relative \(L^2\) error of \(0.003\) (\Cref{tab:params}), an order of magnitude below either posterior-mean error. These scores assess complementary properties of the ensemble; near-unit SSR alone does not establish posterior accuracy. The recorded evaluation times of \(22.3\) and \(227.7\) seconds per scenario are exploratory because the hardware records are not matched (\Cref{app:hardware}).

\begin{table}[H]
\centering
\caption{Posterior inference on the \(64\times64\) Navier--Stokes benchmark from \(64\) noisy observations. CRPS and relative \(L^2\) error assess predictive accuracy, CE assesses marginal calibration, and SSR compares spread with reconstruction error. \(\dagger\) Recorded times include scoring; unmatched hardware prevents a controlled speed comparison (\Cref{app:hardware}).}
\label{tab:e4ns}
\small
\begin{tabular}{lccccr}
\toprule
Method & CRPS $\downarrow$ & SSR $\to 1$ & CE $\downarrow$ & rel.\ $L^2$ $\downarrow$ & s/case$^{\dagger}$ \\
\midrule
FAPS & $\mathbf{0.0121}_{\pm 0.0001}$ & $1.136_{\pm 0.014}$ & $0.0352_{\pm 0.0021}$ & $\mathbf{0.0315}_{\pm 0.0003}$ & 227.7 \\
LAPS (ours) & $0.0158_{\pm 0.0003}$ & $\mathbf{0.989}_{\pm 0.012}$ & $\mathbf{0.0047}_{\pm 0.0011}$ & $0.0369_{\pm 0.0007}$ & 22.3 \\
\bottomrule
\end{tabular}

\end{table}

\begin{figure}[H]
\centering
\includegraphics[width=\linewidth]{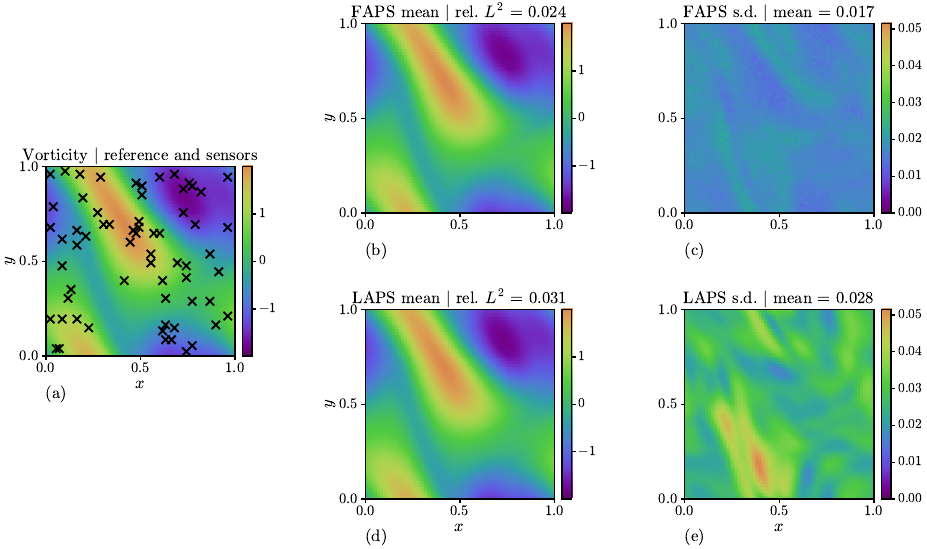}
\caption{Spatial posterior comparison for a representative Navier--Stokes test case. (a) Reference vorticity and the \(64\) observation locations. (b,c) FAPS posterior mean and standard deviation. (d,e) LAPS posterior mean and standard deviation. All estimates use \(128\) posterior draws; posterior-mean panels share one color scale and posterior-standard-deviation panels another.}
\label{fig:ns-spatial}
\end{figure}

\Cref{fig:ns-spatial} shows the test scenario whose LAPS relative \(L^2\) error is the median over the \(100\) cases. Despite observing only \(1.6\%\) of the field, both posterior means recover the dominant large-scale vorticity structures. The FAPS mean follows the realized field more closely, consistent with its lower aggregate reconstruction error in \Cref{tab:e4ns}. The uncertainty fields differ more visibly: in this scenario LAPS assigns a larger and more spatially structured posterior spread than FAPS. Whether a spread is adequate depends on the error it accompanies, which is examined next.

\begin{figure}[H]
\centering
\includegraphics[width=\linewidth]{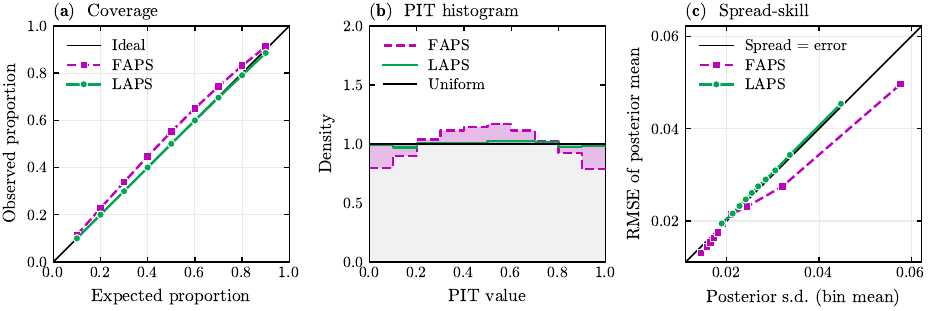}
\caption{Calibration diagnostics for the Navier--Stokes benchmark, pooled over the five training seeds. (a) Empirical versus nominal central-interval coverage; bands show one standard deviation over seeds. (b) Probability integral transform histogram, with uniformity indicating calibrated marginal uncertainty. (c) Spread--skill relationship; the diagonal denotes equality between posterior spread and reconstruction error.}
\label{fig:ns-calib}
\end{figure}

The calibration difference is examined over the full test set in \Cref{fig:ns-calib}. The LAPS coverage curve remains close to the nominal diagonal, whereas FAPS shows systematic over-coverage. The PIT histogram is also closer to uniform for LAPS, while the FAPS histogram is concentrated toward the center. The spread--skill relationship is consistent with the pooled SSR values. These diagnostics assess marginal uncertainty averaged over scenarios and spatial points; spatial points are correlated and are not independent replicates. They do not establish the joint conditional field posterior. A prior-only control separates calibration from the gain due to conditioning: decoded draws of the learned prior, scored identically, give CRPS \(0.302\), CE \(0.064\) and SSR \(0.877\) (\Cref{app:prior-only}), so neither the accuracy nor the calibration of the conditioned ensemble is inherited from the prior.
\FloatBarrier
\FloatBarrier

\subsection{Resolution Dependence}
\label{sec:resolution}

\Cref{sec:analysis-cost} predicts that refining the output grid changes the final decoding and leaves the repeated conditioning unchanged. We examine this with the sampler configuration fixed. In two dimensions the sensors keep their physical locations, and the same latent ensemble is decoded on grids from \(64^2\) to \(512^2\). In one dimension the \(n=512\) runs repeat the sampling with the observation locations mapped to the finer grid (\Cref{app:benchmark-construction}), so they also change the observation operator.

\begin{figure}[t]
\centering
\includegraphics[width=0.46\linewidth]{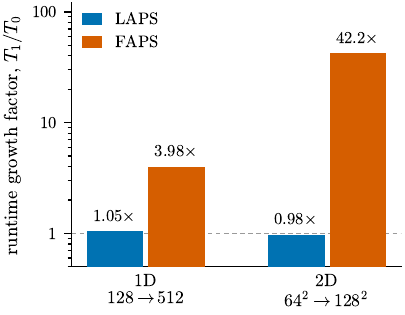}
\caption{Recorded evaluation-time ratios for \(128\rightarrow512\) points in one dimension and \(64^2\rightarrow128^2\) in two dimensions. Ratios are within each implementation. The two-dimensional FAPS bar uses a two-scenario refined-grid probe and an unmatched reference protocol; it is exploratory (\Cref{app:hardware}).}
\label{fig:runtime-growth}
\end{figure}

On Navier--Stokes, conditioning takes \(20.6\)\,s per scenario whatever the output grid, and decoding the \(128\) draws takes \(0.01\)\,s at \(64^2\) and at most \(0.16\)\,s on the finer grids, below \(1\%\) of the conditioning time. The recorded evaluation time, which also contains scoring, stays within \(3\%\) of its \(64^2\) value up to \(512^2\). In one dimension it grows by a factor of \(1.05\) from \(128\) to \(512\) points for LAPS and by \(3.98\) for the released FAPS implementation (\Cref{fig:runtime-growth}). The FAPS \(128^2\) result is a two-scenario probe rather than a matched repeat of the \(64^2\) experiment, so its approximately \(42.2\)-fold recorded increase indicates implementation cost and is not a controlled scaling estimate. Its dense covariance representation exceeds the available \(8\)\,GB device memory at higher resolutions. These measurements are consistent with \eqref{eq:cost-split}. They establish neither a resolution-independent end-to-end cost nor accuracy beyond the learned field distribution, and the cross-method ratios are not speedups. Raw times and the measurement protocol are in \Cref{app:hardware}.

\FloatBarrier

\subsection{Ablation Study}
\label{sec:exp-ablation}

The local covariance property of the VP bridge does not by itself establish better posterior sampling. We examine in turn the local mechanism, the effect of the coefficients on posterior accuracy and the role of multilevel annealing. All posterior ablations use the same trained models, five training seeds, $100$ test scenarios and $128$ posterior draws per scenario. Step sizes are tuned for each variant on held-out scenarios and fixed during testing.

\begin{figure}[t]
\centering
\includegraphics[width=\linewidth]{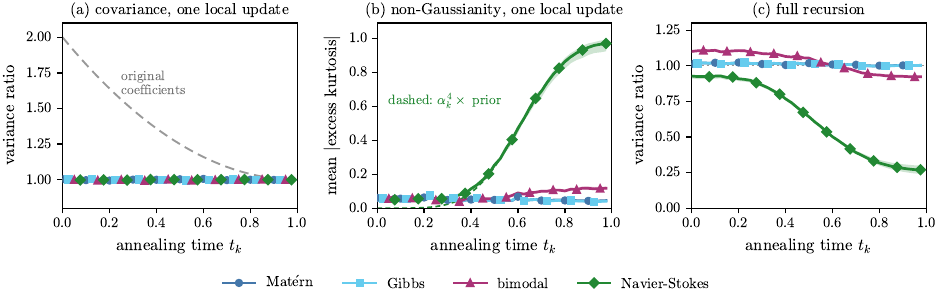}
\caption{Likelihood-free evaluation of the VP bridge with $K=40$, averaged over five training seeds; shaded bands indicate the seed ranges. (a,b) Single local updates using centered learned-prior anchors, matched estimated covariance $\Sigma_\theta$ and unfloored VP coefficients, with $5000$ prior draws per benchmark and seed. (a) Covariance-trace ratio relative to the learned prior; the dashed line gives the original-coefficient factor $1+(1-t_k)^2$. (b) Mean absolute excess kurtosis of $\Sigma_\theta$-whitened coordinates; the dashed line indicates the $\alpha_k^4$ scaling for Navier--Stokes. (c) Complete annealing recursion using learned transport, the implemented $\Sigma_z$, residual noise and the width floor, with $2000$ particles. All panels use $S=200$ Euler steps. The matched conditions in (a,b) differ from those of the complete recursion in (c).}
\label{fig:vp-verification}
\end{figure}

\paragraph{Local covariance and distributional effects}
We first examine the VP bridge without likelihood correction, so that the terminal target is the learned prior. Under matched covariance conditions, \Cref{thm:covariance-preservation} predicts that a local VP update preserves the prior covariance, whereas the original coefficients introduce an additional variance contribution. \Cref{fig:vp-verification}(a) confirms this distinction across all four learned priors. The VP covariance-trace ratio remains within $0.8\%$ of one, while the original coefficients produce the predicted inflation factor $1+(1-t_k)^2$.

Covariance preservation does not imply invariance of the full distribution. For the non-Gaussian Navier--Stokes prior, \Cref{fig:vp-verification}(b) shows a reduction in excess kurtosis consistent with the fourth-order scaling in \eqref{eq:cumulant-contraction}. The distinction becomes more pronounced when the local update is embedded in the complete recursion. With learned transport, the implemented covariance, residual noise and the width floor, the Gaussian covariance traces remain within $0.5\%$ of their prior values, whereas the bimodal and Navier--Stokes traces decrease to approximately $0.92$ and $0.27\pm0.03$, respectively (\Cref{fig:vp-verification}(c)). On the Gaussian benchmarks the latent prior is close to Gaussian (\Cref{fig:vp-verification}(b)), the case in which the likelihood-free recursion returns the prior under exact transport (\Cref{rem:gaussian-closed}). On Navier--Stokes the loss is caused neither by the width floor, nor by the kernel covariance, nor by the transport discretization, and a large part of it remains when the local draw is replaced by the anchor, so that only learned transport and re-bridging act (\Cref{app:null-controls}). The local identity concerns a single update and is not contradicted by these deviations: the VP construction controls local covariance and does not preserve the learned prior throughout annealing.

\paragraph{Effect of VP coefficients}
The two variants differ only in the coefficients and share the width floor and the trained models. As shown in \Cref{tab:ablation-bridge}, VP reduces SWD in every training seed on the three exact-reference benchmarks, by approximately $24\%$ on the two Gaussian benchmarks and $10\%$ on the bimodal benchmark. The Gaussian benchmarks show the mechanism directly: with the original coefficients the ensembles are about $10\%$ too wide pointwise, in the direction of the local inflation of \Cref{sec:analysis-bridge}, whereas with VP they are about $3\%$ too narrow, and the error of the pointwise standard deviation falls by more than half. On the bimodal benchmark, both coefficient choices recover the two configurations, and VP leaves less probability mass between them and lowers the mode-weight error. All distances to the exact posterior improve in every seed (\Cref{tab:ablation-bridge-full}).

On Navier--Stokes the effect is selective. VP brings the spread--skill ratio closer to one in every seed and reduces the mean calibration error, although the confidence interval for the calibration improvement includes zero. At the same time, CRPS increases by approximately $1.2\%$ in every seed, and the relative $L^2$ error does not change beyond the paired interval. Thus, the improvement in covariance control is reflected in posterior spread and exact-reference distances, but it does not translate into uniformly better predictive accuracy.

\begin{table}[t]
\centering
\small
\caption{Original versus VP bridge coefficients on the same trained models and reported annealing schedules. Results are means $\pm$ standard deviations over five training seeds and $100$ test scenarios, with $128$ posterior draws per scenario. The VP configuration corresponds to the reported LAPS results. Differences are paired by training seed, with $95\%$ $t$-intervals; ``Seeds'' counts improvements under VP. For the rows marked $\ast$ (the pointwise width ratio and SSR), differences and improvement counts refer to the distance from one. The width ratio, ModeCov and valley mass are defined in \Cref{app:metrics}.}
\label{tab:ablation-bridge}
\resizebox{\linewidth}{!}{

\begin{tabular}{llcccc}
\toprule
Benchmark & Metric & Original & VP & VP $-$ original [95\% CI] & Seeds \\
\midrule
Mat\'ern & SWD $\downarrow$ & $0.0809_{\pm 0.0011}$ & $0.0614_{\pm 0.0007}$ & $-0.0195$ \,[$-0.0204$,\,$-0.0186$] & 5/5 \\
 & $\mathrm{MAE}_s$ $\downarrow$ & $0.0356_{\pm 0.0009}$ & $0.0133_{\pm 0.0005}$ & $-0.0224$ \,[$-0.0232$,\,$-0.0216$] & 5/5 \\
 & width $\to1$ & $1.108_{\pm 0.003}$ & $0.972_{\pm 0.002}$ & $-0.080$ \,[$-0.086$,\,$-0.074$]$^{\ast}$ & 5/5 \\
\addlinespace[0.25em]
Gibbs & SWD $\downarrow$ & $0.1008_{\pm 0.0009}$ & $0.0762_{\pm 0.0006}$ & $-0.0245$ \,[$-0.0255$,\,$-0.0236$] & 5/5 \\
 & $\mathrm{MAE}_s$ $\downarrow$ & $0.0395_{\pm 0.0005}$ & $0.0147_{\pm 0.0004}$ & $-0.0248$ \,[$-0.0252$,\,$-0.0245$] & 5/5 \\
 & width $\to1$ & $1.090_{\pm 0.003}$ & $0.972_{\pm 0.002}$ & $-0.062$ \,[$-0.067$,\,$-0.056$]$^{\ast}$ & 5/5 \\
\addlinespace[0.25em]
Bimodal & SWD $\downarrow$ & $0.1658_{\pm 0.0033}$ & $0.1499_{\pm 0.0090}$ & $-0.0159$ \,[$-0.0268$,\,$-0.0050$] & 5/5 \\
 & ModeW-MAE $\downarrow$ & $0.0782_{\pm 0.0038}$ & $0.0661_{\pm 0.0071}$ & $-0.0121$ \,[$-0.0209$,\,$-0.0032$] & 5/5 \\
 & ModeCov $\uparrow$ & $1.000_{\pm 0.000}$ & $1.000_{\pm 0.000}$ & $+0.000$ \,[$+0.000$,\,$+0.000$] & tie \\
 & Valley $\to0.04$ & $0.057_{\pm 0.002}$ & $0.045_{\pm 0.001}$ & $-0.012$ \,[$-0.014$,\,$-0.009$] & 5/5 \\
\addlinespace[0.25em]
Navier--Stokes & CE $\downarrow$ & $0.0095_{\pm 0.0034}$ & $0.0047_{\pm 0.0011}$ & $-0.0048$ \,[$-0.0101$,\,$+0.0005$] & 4/5 \\
 & SSR $\to1$ & $1.057_{\pm 0.023}$ & $0.989_{\pm 0.012}$ & $-0.044$ \,[$-0.081$,\,$-0.007$]$^{\ast}$ & 5/5 \\
 & CRPS $\downarrow$ & $0.0156_{\pm 0.0003}$ & $0.0158_{\pm 0.0003}$ & $+0.0002$ \,[$+0.0001$,\,$+0.0003$] & 0/5 \\
 & rel.\ $L^2$ $\downarrow$ & $0.0367_{\pm 0.0008}$ & $0.0369_{\pm 0.0007}$ & $+0.0002$ \,[$-0.0001$,\,$+0.0004$] & 1/5 \\
\bottomrule
\end{tabular}
}
\end{table}

\paragraph{Role of multilevel annealing}
We finally vary the number of annealing levels at a fixed total MALA-update budget $N_{\rm MALA}=KL$. At $K=1$, the initial coefficient satisfies $\alpha_0=0$, and the local target reduces to
\begin{equation}
    \pi_0^{\mathrm{loc}}(\mathrm d\vz)
    \propto
    \exp[-\Phi_\psi(\vz;\vy)]
    \N(\mathrm d\vz;\vzero,\Sigma_z).
\end{equation}
The single-level configuration therefore replaces the learned latent prior with a Gaussian approximation. This comparison evaluates the combined effect of the multilevel construction and retention of the learned prior, rather than annealing depth alone.

\Cref{tab:ablation-schedule} shows that the Gaussian replacement substantially degrades posterior approximation. On the bimodal benchmark, it produces a broad ensemble with approximately $31\%$ of its samples between the two configurations, compared with $4.5\%$ under the reported $K=40$ schedule. Its SWD is approximately one-third larger, although its posterior mean is more accurate. On Navier--Stokes, the single-level configuration more than doubles CRPS, while its spread--skill ratio remains close to that of the reported sampler. Intermediate schedules generally improve upon the Gaussian replacement, but increasing the number of levels does not guarantee monotonic improvement: on the bimodal benchmark, $K=10$ achieves a slightly smaller SWD than $K=40$. These results support the multilevel construction as a means of retaining information from the learned prior, without isolating a universal benefit from additional annealing levels.

A complementary comparison uses direct MCMC under the same learned latent prior, avoiding the Gaussian replacement inherent in the single-level ablation. At approximately matched wall-clock budgets, direct pCN achieves SWD comparable to LAPS on the Gaussian benchmarks and slightly better SWD on the bimodal benchmark. LAPS provides substantially better marginal calibration on Navier--Stokes at the tested budget. Direct MCMC is run at a fixed budget and need not have converged, so the comparison establishes neither a general advantage for annealing nor superiority over converged MCMC. The comparison conditions and additional results are provided in \Cref{app:direct-mcmc}.

\begin{table}[t]
\centering
\small
\caption{Effect of the annealing schedule at fixed MALA-update budgets, $N_{\rm MALA}=KL=2000$ for the bimodal benchmark and $8000$ for Navier--Stokes. Results are means $\pm$ standard deviations over five training seeds and $100$ test scenarios. The reported configurations use $(K,L)=(40,50)$ and $(80,100)$, respectively. Transport work and wall-clock costs are not matched. The $K=1$ configuration replaces the learned latent prior with a Gaussian distribution. Valley mass is defined in \Cref{app:metrics}.}
\label{tab:ablation-schedule}
\resizebox{\linewidth}{!}{

\begin{tabular}{lcccccc}
\toprule
 & \multicolumn{3}{c}{Bimodal} & \multicolumn{3}{c}{Navier--Stokes} \\
\cmidrule(lr){2-4}\cmidrule(lr){5-7}
$(K,L)$ & SWD $\downarrow$ & $\mathrm{RMSE}_m$ $\downarrow$ & Valley $\to0.04$ & CRPS $\downarrow$ & CE $\downarrow$ & SSR $\to1$ \\
\midrule
reported & $0.1499_{\pm 0.0090}$ & $0.0678_{\pm 0.0056}$ & $0.045_{\pm 0.001}$ & $0.0158_{\pm 0.0003}$ & $0.0047_{\pm 0.0011}$ & $0.989_{\pm 0.012}$ \\
$(10,\,N_{\rm MALA}/10)$ & $0.1410_{\pm 0.0061}$ & $0.0615_{\pm 0.0028}$ & $0.066_{\pm 0.007}$ & $0.0198_{\pm 0.0001}$ & $0.0155_{\pm 0.0010}$ & $1.004_{\pm 0.004}$ \\
$(1,\,N_{\rm MALA})$ & $0.2018_{\pm 0.0005}$ & $0.0386_{\pm 0.0023}$ & $0.307_{\pm 0.006}$ & $0.0340_{\pm 0.0001}$ & $0.0196_{\pm 0.0009}$ & $0.988_{\pm 0.003}$ \\
\bottomrule
\end{tabular}
}
\end{table}
\section{Discussion, Limitations, and Conclusion}
\label{sec:discussion}
\subsection{Discussion and Limitations}
The evidence supports the computational separation, with a measurable accuracy--calibration trade-off: on Navier--Stokes, FAPS gives lower CRPS and posterior-mean error and LAPS better aggregate marginal calibration. The prior-only control shows that this calibration is produced by conditioning and not inherited from the prior; marginal calibration still cannot establish the joint posterior. On the bimodal benchmark both configurations are retained, but their weights are flattened toward one half. The resolution sweep agrees with the cost analysis: repeated conditioning is independent of the requested output grid, whereas final decoding remains linear in its size. Unmatched devices and the short FAPS refined-grid probe limit cross-method speed and scaling claims.

The VP bridge is a local mechanism. It lowers exact-reference distances and brings spread closer to nominal in every seed, yet it raises Navier--Stokes CRPS slightly, and its covariance identity does not extend to the complete recursion: in the likelihood-free Navier--Stokes diagnostic the trace falls to about \(0.27\) of its prior value, a loss that does not depend on the width floor and is already substantial when only learned transport and re-bridging act (\Cref{app:null-controls}). The calibration observed on Navier--Stokes is therefore an empirical property of the sampler and does not indicate that the complete recursion preserves the learned prior. Direct MCMC under the same prior is comparably accurate on the one-dimensional benchmarks, so the schedule and direct-MCMC controls support the reported configuration at the tested budgets without identifying annealing as the cause of the cross-method gains.

The target posterior is that of the learned latent prior and decoder. Its coherence does not remove compression error, missing rare modes or training bias, and new coordinates cannot recover field structure absent from that model. The implemented local Gaussian targets, residual noise, coefficient floor and finite-step transport approximate the ideal conditional transitions, and Metropolis correction targets each local distribution without a finite-step mixing guarantee. Complete-method comparisons combine representation, bridge and budget choices, and the two-dimensional evidence comes from a single benchmark. Timings were not recorded on matched devices, so no cross-method speedup is claimed.

\subsection{Conclusion}
\label{sec:conclusion}

We asked whether repeated posterior conditioning of a sparsely observed random field can be carried out in a fixed-dimensional latent space, separately from the output grid, without losing a credible posterior. The separation holds by construction: a coherent coordinate-decoded field model confines every repeated operation to the latent state and the sensors, and the recorded LAPS evaluation time changes by less than \(3\%\) over a \(64\)-fold change of output size. Under the learned model, the posterior approximation has empirical support. Errors against exact Gaussian posteriors are close to the exact-draw reference, both configurations of a bimodal posterior are retained, with mode weights biased toward one half, and Navier--Stokes ensembles are better calibrated than those of the field-space baseline, at higher CRPS and field error. Exactness is limited by the learned representation and by approximate, finite-step annealed transitions; within them the VP bridge controls local covariance inflation, and guarantees neither full-recursion invariance nor uniformly better accuracy.

\appendix
\renewcommand{\thesection}{\Alph{section}}
\renewcommand{\theHtable}{\thesection.\arabic{table}}
\renewcommand{\theHfigure}{\thesection.\arabic{figure}}

\appendixsection{Proofs of Theoretical Results}
\label{app:proofs}
\appendixsubsection{Coherent Field Prior and Posterior}
\label{app:process}

\begin{proof}[Proof of \Cref{thm:process-consistency}]
The potential $\Phi(\cdot\,;\vy)$ is continuous on $C(\X)$, so $\Phi_\psi=\Phi(D_\psi(\cdot);\vy)$ is Borel measurable, and $0<e^{-\Phi_\psi}\le1$ gives $0<\mathcal Z_\theta(\vy)\le1$; the latent posterior $\pi_\theta^{\vy}$ is therefore well defined. Since $D_\psi:\R^d\to C(\X)$ is Borel measurable, the pushforward measures $\widehat\mu=(D_\psi)_{\#}\nu_\theta$ and $\widehat\pi^{\vy}=(D_\psi)_{\#}\pi_\theta^{\vy}$ are Borel probability measures on $C(\X)$.

For any $n$ spatial locations $X_n=(x_1,\ldots,x_n)$, let $P_{X_n}:C(\X)\to\R^n$ denote evaluation at these locations. By construction, $P_{X_n}\circ D_\psi=D_\psi(\cdot,X_n)$, so
\[
\widehat\mu_{X_n}=(P_{X_n}\circ D_\psi)_{\#}\nu_\theta,
\qquad
\widehat\pi_{X_n}^{\vy}=(P_{X_n}\circ D_\psi)_{\#}\pi_\theta^{\vy}.
\]
Permuting or restricting the locations in $X_n$ corresponds to applying the same operation to $P_{X_n}$. The finite-dimensional distributions therefore satisfy the required consistency relations. Since $\X$ is compact metric, $C(\X)$ is separable, and its Borel $\sigma$-algebra is generated by evaluations on a countable dense subset of $\X$. The cylinder sets defined by finitely many evaluations form a $\pi$-system that generates this $\sigma$-algebra, so two Borel probability measures with the same finite-dimensional distributions coincide.

It remains to establish the posterior identity. The definition of $\Phi_\psi$ and the pushforward integration formula give
\begin{equation}
\begin{aligned}
\mathcal Z_\theta(\vy)
&=\int_{\R^d}e^{-\Phi_\psi(\vz;\vy)}\,\nu_\theta(\mathrm d\vz)\\
&=\int_{C(\X)}e^{-\Phi(u;\vy)}\,\widehat\mu(\mathrm du).
\end{aligned}
\end{equation}
For any Borel set $B\subseteq C(\X)$, latent Bayesian conditioning gives
\begin{equation}
\begin{aligned}
\widehat\pi^{\vy}(B)
&=\pi_\theta^{\vy}\bigl(D_\psi^{-1}(B)\bigr)\\
&=\frac{1}{\mathcal Z_\theta(\vy)}
\int_{D_\psi^{-1}(B)}
e^{-\Phi_\psi(\vz;\vy)}\,\nu_\theta(\mathrm d\vz)\\
&=\frac{1}{\mathcal Z_\theta(\vy)}
\int_B e^{-\Phi(u;\vy)}\,\widehat\mu(\mathrm du).
\end{aligned}
\end{equation}
This establishes \eqref{eq:decoded-field-posterior} without requiring $D_\psi$ to be injective.
\end{proof}

\appendixsubsection{Ideal Annealing and Posterior Sampling}
\label{app:ideal-transition}

\begin{proof}[Proof of \Cref{prop:ideal-transition}]
Following the conditional sampling principle of \citet[Proposition~1]{shi2026faps}, consider the joint probability measure
\begin{equation}
\Pi_{t'}(\mathrm d\vz_{t'},\mathrm d\vz)
=
Q_{t'}(\mathrm d\vz_{t'}\mid\vz)\,
\pi_\theta^{\vy}(\mathrm d\vz).
\end{equation}
Its marginals are $\pi_{t'}^{\vy}$ and $\pi_\theta^{\vy}$ by the definition of the annealed distributions. Since $\R^d$ is Polish, a regular conditional distribution of $\vz$ given $\vz_{t'}$ exists.

Drawing $\vz_{t'}\sim\pi_{t'}^{\vy}$ and then $\vz$ from its exact conditional distribution reproduces the joint measure $\Pi_{t'}$. Consequently, the terminal draw has marginal distribution $\pi_\theta^{\vy}$. Applying the bridge kernel at a later time $t>t'$ gives
\begin{equation}
\int_{\R^d}
Q_t(\mathrm d\vz_t\mid\vz)\,
\pi_\theta^{\vy}(\mathrm d\vz)
=
\pi_t^{\vy}(\mathrm d\vz_t).
\end{equation}
The transition therefore maps $\pi_{t'}^{\vy}$ to $\pi_t^{\vy}$. At $t=1$, the Dirac measure $Q_1(\cdot\mid\vz)=\delta_{\vz}$ gives the target distribution $\pi_\theta^{\vy}$. The result requires neither a Gaussian latent prior nor an explicit prior density.
\end{proof}

\noindent\textit{Gaussian reference calculation.}
Consider the special case in which the encoded latent distribution and the learned prior are both $\N(\vzero,I_d)$, $\Sigma_z=I_d$, and $\sigma_{\min}=0$. Assume exact flow matching along the independent-coupling path and use the unfloored VP coefficients. We show that the VP local prior coincides with the conditional prior of the ideal bridge.

For independent standard Gaussian endpoints, let $c_t=t^2+(1-t)^2$. The exact flow-matching velocity and its terminal transport are
\begin{equation}
v_\theta(\vz,t)
=
\frac{2t-1}{c_t}\vz,
\qquad
T^\theta_{t,1}(\vz_t)
=
\frac{\vz_t}{\sqrt{c_t}}.
\end{equation}
Completing the square in the product of the bridge density $Q_t(\vz_t\mid\vz)$ and the standard Gaussian prior gives the conditional distribution
\begin{equation}
\N\!\left(
\mathrm d\vz;
\frac{t}{c_t}\vz_t,
\frac{(1-t)^2}{c_t}I_d
\right).
\label{eq:gaussian-conditional-match}
\end{equation}
For $t=t_k<1$, the VP coefficients reduce to
\[
\alpha_k=\frac{t_k}{\sqrt{c_{t_k}}},
\qquad
\lambda_k=\frac{1-t_k}{\sqrt{c_{t_k}}}.
\]
Since $\bar{\vz}_k=T^\theta_{t_k,1}(\vz_k)=\vz_k/\sqrt{c_{t_k}}$, the local Gaussian prior satisfies
\begin{equation}
\begin{aligned}
q_k(\mathrm d\vz\mid\bar{\vz}_k)
&=\N\!\left(
\mathrm d\vz;
\alpha_k\bar{\vz}_k,
\lambda_k^2I_d
\right)\\
&=\N\!\left(
\mathrm d\vz;
\frac{t_k}{c_{t_k}}\vz_k,
\frac{(1-t_k)^2}{c_{t_k}}I_d
\right).
\end{aligned}
\end{equation}
This agrees with the ideal conditional prior in \eqref{eq:gaussian-conditional-match}. Multiplication by the same observation likelihood gives identical terminal conditional posteriors. With exact local sampling and ideal re-bridging, \Cref{prop:ideal-transition} applies at every level, yielding $\pi_\theta^{\vy}$ at the final level.

\appendixsubsection{Local Covariance Preservation and Mismatch}
\label{app:covariance-preservation}
\label{app:bridge-distribution}

\begin{proof}[Proof of \Cref{thm:covariance-preservation}]
Consider a draw from the VP local prior,
\[
\mathbf w_k
=
\alpha_k\bar{\vz}_k
+
\lambda_k\Sigma_z^{1/2}\veps_k,
\qquad
\veps_k\sim\N(\vzero,I_d),
\]
where $\veps_k$ is independent of $\bar{\vz}_k$. Independence and finite second moments give
\begin{equation}
\operatorname{Cov}(\mathbf w_k)
=
\alpha_k^2\operatorname{Cov}(\bar{\vz}_k)
+
\lambda_k^2\Sigma_z.
\label{eq:vp-covariance}
\end{equation}
Substituting $\operatorname{Cov}(\bar{\vz}_k)=\Sigma_\theta+\Delta_k$, we obtain
\[
\begin{aligned}
\operatorname{Cov}(\mathbf w_k)-\Sigma_\theta
&=\alpha_k^2(\Sigma_\theta+\Delta_k)
+\lambda_k^2\Sigma_z-\Sigma_\theta\\
&=\alpha_k^2\Delta_k
+\lambda_k^2(\Sigma_z-\Sigma_\theta)\\
&\quad+(\alpha_k^2+\lambda_k^2-1)\Sigma_\theta.
\end{aligned}
\]
The unfloored VP coefficients satisfy $\alpha_k^2+\lambda_k^2=1$, so the final term vanishes. This gives \eqref{eq:bridge-covariance-mismatch}. If $\Delta_k=0$ and $\Sigma_z=\Sigma_\theta$, both remaining terms vanish, yielding
\[
\operatorname{Cov}(\mathbf w_k)=\Sigma_\theta,
\]
which establishes \eqref{eq:covariance-preservation}. The argument requires no Gaussian assumption on the learned prior or transported endpoint.
\end{proof}

The corresponding first moment satisfies
\begin{equation}
\E[\mathbf w_k]
=
\alpha_k\E[\bar{\vz}_k].
\end{equation}
Thus matching covariance alone does not guarantee mean preservation. When the learned prior and covariance-matched endpoint are both centered, the local draw preserves their first two moments.

For the implemented coefficients, an active width floor may violate $\alpha_k^2+\lambda_k^2=1$. The covariance error then satisfies
\begin{equation}
\begin{aligned}
\operatorname{Cov}(\mathbf w_k)-\Sigma_\theta
={}&\alpha_k^2\Delta_k
+\lambda_k^2(\Sigma_z-\Sigma_\theta)\\
&+(\alpha_k^2+\lambda_k^2-1)\Sigma_\theta.
\end{aligned}
\end{equation}
The final term accounts for the departure from the exact covariance budget.

For non-Gaussian distributions, covariance preservation does not imply distributional invariance. If the fourth moments exist, independence and the vanishing fourth-order cumulant of Gaussian noise yield
\begin{equation}
\kappa_4(\mathbf w_k)
=
\alpha_k^4\kappa_4(\bar{\vz}_k),
\label{eq:cumulant-contraction}
\end{equation}
where $\kappa_4$ denotes the fourth-order cumulant tensor. Hence the Gaussian perturbation can change higher-order structure even when covariance is preserved. These identities apply to a single local prior draw and do not imply invariance of the complete annealing recursion.

\appendixsection{Experimental Protocol}
\label{app:protocol}

\appendixsubsection{Benchmarks and Observations}
\label{app:benchmark-construction}

We consider two Gaussian-process benchmarks, a bimodal random-field model and a two-dimensional Navier--Stokes dataset. The first three admit exact posterior references, while the Navier--Stokes benchmark is evaluated through predictive accuracy and uncertainty calibration. The benchmark configurations are summarized in \Cref{tab:benchmark-settings}.

\begin{table*}[t]
\centering
\small
\setlength{\tabcolsep}{4pt}
\caption{Benchmark configurations. Each benchmark uses $100$ test scenarios. Training and validation sample counts are reported together.}
\label{tab:benchmark-settings}
\begin{tabular*}{\textwidth}{@{\extracolsep{\fill}}lccccc@{}}
\toprule
Benchmark & Grid & Train / Val. & Observations & Noise $\sigma$ & Exact posterior \\
\midrule
Mat\'ern
    & $128$ & $20{,}000/1{,}000$ & $7$ random & $0.1$ & Gaussian \\
Gibbs
    & $128$ & $20{,}000/1{,}000$ & $7$ random & $0.1$ & Gaussian \\
Bimodal
    & $128$ & $20{,}000/1{,}000$ & $5$ fixed & $0.05$ & Gaussian mixture \\
Navier--Stokes
    & $64\times64$ & $30{,}000/2{,}000$ & $64$ random & $0.0434$ & None \\
\bottomrule
\end{tabular*}
\end{table*}

\paragraph{Gaussian-process benchmarks}
Following \citet{shi2026faps}, the Mat\'ern and Gibbs random fields are defined on $\X=[0,1]$ with zero mean and unit marginal variance. The stationary Mat\'ern-$3/2$ covariance is
\begin{equation}
    k_{\mathrm m}(x,x')
    =
    \left(1+\frac{\sqrt{3}h}{\ell_{\mathrm m}}\right)
    \exp\!\left(-\frac{\sqrt{3}h}{\ell_{\mathrm m}}\right),
    \qquad
    h=|x-x'|,
    \quad
    \ell_{\mathrm m}=0.3.
\end{equation}
The non-stationary Gibbs covariance is
\begin{equation}
    k_{\mathrm g}(x,x')
    =
    \sqrt{
        \frac{2\ell(x)\ell(x')}
        {\ell^2(x)+\ell^2(x')}
    }
    \exp\!\left[
        -\frac{(x-x')^2}
        {\ell^2(x)+\ell^2(x')}
    \right],
    \qquad
    \ell(x)=0.05+0.25x.
\end{equation}

For each test scenario, seven distinct observation locations are sampled uniformly from the training grid. Exact posterior means and covariances are obtained by Gaussian conditioning, with $128$ independent posterior draws used as the reference. Evaluation uses $128$ and $512$ spatial points. At the finer resolution, the observed values are retained while their locations are mapped to the nearest evaluation-grid nodes; the corresponding likelihood and exact posterior use the mapped observation operator, and the sampler is run again. The \(n=512\) results are therefore not a re-decoding of the \(n=128\) ensembles. Over the five training seeds, LAPS attains an SWD of \(0.0613\pm0.0006\) on Mat\'ern and \(0.0778\pm0.0019\) on Gibbs at \(n=512\), against \(0.0614\pm0.0007\) and \(0.0767\pm0.0014\) for independent ensembles of exact draws. Twenty additional scenarios are reserved for step-size tuning, and data generation uses seed $2026$.

\paragraph{Bimodal benchmark}
The bimodal random field combines a localized Gaussian bump with an independent Gaussian background:
\begin{equation}
    u(x)
    =
    a_b\exp\!\left[
        -\frac{(x-x_c)^2}{2w_b^2}
    \right]
    +b(x),
    \qquad
    x_c\sim
    \frac12\delta_{0.35}
    +
    \frac12\delta_{0.65},
\end{equation}
where $a_b=2.0$, $w_b=0.08$, and $b$ is a zero-mean Gaussian process with covariance
\begin{equation}
    k_b(x,x')
    =
    \sigma_b^2
    \exp\!\left[
        -\frac{(x-x')^2}{2\ell_b^2}
    \right],
    \qquad
    \sigma_b^2=0.2,
    \quad
    \ell_b=0.15.
\end{equation}

Observations are collected at the grid nodes nearest to $0.05$, $0.18$, $0.50$, $0.82$ and $0.95$. Conditional on either bump center, the posterior is Gaussian, and the component weights are obtained from the corresponding observation likelihoods. The exact posterior is therefore a two-component Gaussian mixture, from which $128$ independent reference draws are generated per scenario. Both component weights exceed $0.05$ in all test scenarios. The benchmark uses $20$ additional tuning scenarios and data-generation seed $2026$.

\paragraph{Navier--Stokes benchmark}
\label{app:ns-data}
The two-dimensional benchmark uses the $64\times64$ vorticity snapshots on the periodic torus $\mathbb T^2$ of the MINO release \citep{shi2025mino}, which are drawn from the Navier--Stokes dataset of \citet{li2021fno} following the preprocessing of \citet{kerrigan2024ffm} and \citet{shi2025ofm}. The fields are used without additional preprocessing. For each scenario, $64$ observation locations are sampled uniformly without replacement, with Gaussian noise of standard deviation $\sigma=0.05\times0.867\approx0.0434$. Five additional scenarios are reserved for step-size tuning, and the observation locations and noise are generated using seed $4046$.

The benchmark follows the released data ordering for its training, validation and test splits. We use the released fields directly and refer to these works for the simulation set-up.

\appendixsubsection{Evaluation Metrics}
\label{app:evaluation}
\label{app:metrics}

All metrics are computed using a common implementation. For each scenario, let $\{\vu^{(i)}\}_{i=1}^{N}$ denote $N$ posterior samples evaluated at $n$ spatial points, $\{\vu_\pi^{(i)}\}_{i=1}^{N}$ an independent reference ensemble when the exact posterior is available, and $\vu_{\rm true}$ the realized field. The sample mean and pointwise standard deviation are denoted by $\widehat m$ and $\widehat s$, while $m$ and $s$ denote their exact posterior counterparts.

\paragraph{Distributional accuracy}
The sliced Wasserstein distance (SWD) measures the discrepancy between two distributions by averaging one-dimensional $2$-Wasserstein distances \citep{villani2009ot} over random unit projections. We use $256$ directions and average the estimates over $10$ independent direction sets. Maximum mean discrepancy (MMD) measures distributional differences through a reproducing kernel Hilbert space. We use the unbiased squared-MMD estimator with the Gaussian RBF kernel
\[
    k(\vu,\mathbf{v})
    =
    \exp\!\left(
        -\frac{\lVert\vu-\mathbf{v}\rVert^2}{n}
    \right),
\]
where $n$ is the number of evaluation points, and report $\sqrt{\max(\widehat{\mathrm{MMD}}^2,0)}$. Both metrics follow the implementation of \citet{shi2026faps}, with smaller values indicating closer agreement between sampled and reference distributions.

The posterior mean and pointwise standard deviation are assessed using
\begin{equation}
\begin{aligned}
    \mathrm{RMSE}_m
    &=
    \left[
        \frac1n\sum_{j=1}^{n}
        \bigl(\widehat m(x_j)-m(x_j)\bigr)^2
    \right]^{1/2},\\
    \mathrm{MAE}_s
    &=
    \frac1n\sum_{j=1}^{n}
    \left|
        \widehat s(x_j)-s(x_j)
    \right|.
\end{aligned}
\end{equation}
These metrics measure errors in the posterior mean and pointwise uncertainty, respectively. The width ratio is the spatial average of $\widehat s(x_j)/s(x_j)$; a value of one indicates agreement with the exact posterior standard deviation.

\paragraph{Predictive accuracy}
The continuous ranked probability score (CRPS) measures the agreement between the predictive distribution and the realized field. Writing $u_j^{(i)}$ for the $j$th component of $\vu^{(i)}$, its ensemble estimator is
\begin{equation}
\begin{aligned}
    \mathrm{CRPS}
    =
    \frac1n\sum_{j=1}^{n}
    \Bigg[
        &\frac1N\sum_{i=1}^{N}
        |u_j^{(i)}-u_{{\rm true},j}|\\
        &-\frac{1}{2N^2}
        \sum_{i,\ell=1}^{N}
        |u_j^{(i)}-u_j^{(\ell)}|
    \Bigg].
\end{aligned}
\end{equation}
Smaller CRPS indicates better predictive accuracy. The relative $L^2$ error measures reconstruction accuracy through the normalized difference between the posterior mean and the realized field.

\paragraph{Uncertainty calibration}
The spread--skill ratio (SSR) compares the ensemble spread with the error of the posterior mean. Let $V$ and $E$ denote the averages over scenarios of the pointwise ensemble variance and squared posterior-mean error:
\begin{equation}
\begin{aligned}
    V
    &=
    \frac1n\sum_{j=1}^{n}
    \widehat{\operatorname{Var}}_i\,u_j^{(i)},\\
    E
    &=
    \frac1n\sum_{j=1}^{n}
    \bigl(\widehat m(x_j)-u_{{\rm true},j}\bigr)^2.
\end{aligned}
\end{equation}
The pooled spread--skill ratio is
\begin{equation}
    \mathrm{SSR}=\sqrt{\frac{V}{E}},
\end{equation}
where both quantities are averaged over scenarios before taking their ratio. Values near one indicate agreement between ensemble spread and posterior-mean error.

The calibration error (CE) measures the deviation of empirical interval coverage from its nominal level. Let $\widehat C(\gamma)$ be the empirical coverage of the central $\gamma$-level posterior interval, obtained from the corresponding sample quantiles and averaged over spatial points and scenarios. We define
\begin{equation}
    \mathrm{CE}
    =
    \frac19
    \sum_{\gamma\in\{0.1,\ldots,0.9\}}
    \left|
        \widehat C(\gamma)-\gamma
    \right|.
\end{equation}
Smaller CE indicates better marginal calibration. Probability integral transform (PIT) values provide an additional calibration measure by evaluating the empirical predictive cumulative distribution at the realized field values; the histogram uses the randomized form $(\rho+U)/(N+1)$, where $\rho$ is the number of draws below the realized value and $U$ is uniform on $[0,1]$, which is exactly uniform for a calibrated ensemble of $N$ draws.

\paragraph{Mode metrics}
For the bimodal benchmark, each posterior sample is assigned to one of two template regions according to
\begin{equation}
    \langle\vu,\boldsymbol{\delta}\rangle>0,
\end{equation}
where $\boldsymbol{\delta}$ is the difference between the two bump templates. The same assignment rule is applied to all methods.

Mode coverage (ModeCov) is the proportion of test scenarios in which both regions contain at least $5\%$ of the posterior samples. Mode-weight mean absolute error (ModeW-MAE) is the average absolute difference between the sampled weight of the first region and its exact mixture weight. Since template assignment and mixture-component membership are not identical when the components overlap, ModeW-MAE measures an approximation to the component-weight error.

The valley mass measures the fraction of posterior samples lying between the two configurations. It is defined as the proportion of samples whose least-squares amplitudes on the two bump templates differ by less than $0.4$. Smaller valley mass indicates less probability mass between the modes.

All scenario-level metrics are averaged over the test scenarios. When multiple trained models are evaluated, results are reported as means and standard deviations across training seeds.

\appendixsubsection{Comparison Protocol}
\label{app:baseline}
\label{app:reproduction}

The experiments compare LAPS with the field-space annealing method FAPS \citep{shi2026faps} and three amortized neural-process baselines: CNP, ANP and FlowNP \citep{garnelo2018cnp,kim2019anp,abuhamad2025flownp}. All methods use the same benchmark fields, observation scenarios and evaluation procedures. Each posterior ensemble contains $128$ draws, and results are evaluated over $100$ test scenarios. The learned methods are assessed across five independently trained models.

FAPS is evaluated using the authors' original sampler and published configuration. Released prior models are used where available, with additional priors trained using the authors' code and the same training fields to obtain five training seeds. The neural-process baselines are trained on the corresponding datasets. The network and sampler settings are provided in \Cref{app:parameters}.

\Cref{tab:reproduction} compares our FAPS re-run with the published results on benchmarks for which reference values are available. The re-run closely matches the published Mat\'ern results, with differences below $3\%$. On the Gibbs benchmark, its SWD and MMD are approximately $12$--$17\%$ lower than the published values, so the comparison is conservative with respect to this baseline. The CNP and FlowNP reproductions agree with their published results within $4\%$, whereas ANP achieves better reported scores using a four-sample importance-weighted training objective. All comparisons use the reproduced baselines.

\begin{table}[tbp]
\centering
\small
\setlength{\tabcolsep}{4pt}
\caption{Reproduction of the FAPS results reported by \citet{shi2026faps}. The re-run uses the released prior weights. Differences are relative to the published values; positive values indicate larger errors in the re-run.}
\label{tab:reproduction}
\begin{tabular}{llccr}
\toprule
Benchmark & Metric & Published & This work & Difference \\
\midrule
Mat\'ern, $n=128$
    & SWD & $0.142$ & $0.1405$ & $-1.1\%$ \\
    & MMD & $0.128$ & $0.1314$ & $+2.7\%$ \\
Mat\'ern, $n=512$
    & SWD & $0.147$ & $0.1467$ & $-0.2\%$ \\
    & MMD & $0.136$ & $0.1360$ & $+0.0\%$ \\
\addlinespace[0.3em]
Gibbs, $n=128$
    & SWD & $0.194$ & $0.1706$ & $-12.1\%$ \\
    & MMD & $0.167$ & $0.1392$ & $-16.6\%$ \\
\bottomrule
\end{tabular}
\end{table}

In the paired ablations of \Cref{sec:exp-ablation}, the trained models, observation scenarios, sampler seeds and scoring procedures are held fixed. Step sizes are tuned for each variant on held-out scenarios and frozen before testing, so each variant is evaluated at its own tuned step sizes.

\paragraph{Prior-only control}
\label{app:prior-only}
The control in \Cref{sec:ns} uses no observation. For each training seed and test scenario, $128$ draws of the learned prior are generated with $S=20$ Euler steps, decoded on the $64\times64$ grid and scored with the code of \Cref{tab:e4ns}. Over the five training seeds this gives CRPS $0.3022\pm0.0007$, relative $L^2$ error $0.610\pm0.002$, CE $0.0639\pm0.0031$ and SSR $0.877\pm0.007$.

\paragraph{Matched-update control}
\label{app:matched-budget}
On Navier--Stokes the reported LAPS configuration uses $8000$ local updates per scenario and FAPS $2000$. With $(K,L)=(40,50)$, that is, $2000$ updates with step sizes tuned for this schedule and otherwise the reported configuration, LAPS gives CRPS $0.0162\pm0.0002$, relative $L^2$ error $0.0376\pm0.0005$, CE $0.0115\pm0.0014$ and SSR $1.021\pm0.007$ over the five training seeds. The ordering of the two methods in \Cref{tab:e4ns} is unchanged in every column.

\paragraph{Likelihood-free recursion controls}
\label{app:null-controls}
The Navier--Stokes recursion of \Cref{fig:vp-verification}(c) is repeated with one ingredient changed at a time, at $K=40$ with $2000$ particles and five training seeds. The terminal trace ratio is $0.27\pm0.03$ as implemented, $0.27\pm0.03$ without the width floor, $0.26\pm0.03$ with the $S=20$ transport of the posterior runs, and $0.23\pm0.02$ when the kernel uses $\Sigma_\theta$ in place of $\Sigma_z$. When the local draw is replaced by the anchor itself ($\alpha_k=1$, $\lambda_k=0$), so that only learned transport and re-bridging act, the ratio is $0.56\pm0.02$. With exact transport in the sense of \Cref{rem:gaussian-closed} this last recursion returns the prior whatever its law, so its loss measures the error of the learned transport on re-bridged states.

\paragraph{Runtime measurements}
\label{app:hardware}
Evaluation time is measured as the wall clock of the evaluation loop divided by the number of scenarios. It includes posterior sampling, field decoding and metric computation, but excludes checkpoint loading, data loading and offline training. The usual evaluation contains $100$ scenarios, with shorter probes identified separately in the resolution experiment. \Cref{tab:timings} lists the recorded times behind \Cref{sec:resolution}.

LAPS evaluation timings were recorded on an NVIDIA GeForce RTX 4060 with $8$\,GB device memory, using PyTorch 2.6.0 and CUDA 12.6. The one-dimensional FAPS and neural-process timings were recorded on an NVIDIA RTX 2000 Ada Generation Laptop GPU, while the GPU model used for the Navier--Stokes FAPS timing was not recorded. The resulting cross-method timing measurements are not controlled hardware comparisons and are therefore not interpreted as established speedups. Within-method resolution comparisons retain their own recorded configurations, with the limitations of short refined-grid probes stated in \Cref{sec:resolution}. The direct-MCMC comparison uses a separate matched-process timing protocol described in \Cref{app:direct-mcmc}.

The split of \eqref{eq:cost-split} is measured separately, in one process on the RTX 4060, for Navier--Stokes training seed $0$ and five test scenarios after one warm-up scenario. Conditioning, which comprises the likelihood and preconditioner set-up and the annealed recursion, takes $20.6\pm0.5$\,s per scenario. Decoding the $128$ draws, including the trunk evaluation at the output locations, takes $0.009$, $0.043$, $0.16$ and $0.12$\,s at $64^2$, $128^2$, $256^2$ and $512^2$.

\begin{table}[tbp]
\centering
\small
\caption{Recorded evaluation time per scenario (s), including sampling, decoding and scoring. One-dimensional entries are those of the Mat\'ern benchmark, averaged over five training seeds; the LAPS Navier--Stokes sweep uses training seed $0$ with the sampler configuration fixed. The FAPS $64^2$ entry averages five sampler seeds and its $128^2$ entry is a two-scenario probe. Devices are not matched across methods. A dash denotes a setting that was not run.}
\label{tab:timings}
\begin{tabular}{llrr}
\toprule
Benchmark & Output grid & LAPS & FAPS \\
\midrule
Mat\'ern & $128$ & $5.49$ & $12.52$ \\
Mat\'ern & $512$ & $5.75$ & $49.87$ \\
Navier--Stokes & $64^2$ & $22.3$ & $227.7$ \\
Navier--Stokes & $128^2$ & $21.8$ & $9609.5$ \\
Navier--Stokes & $256^2$ & $21.8$ & -- \\
Navier--Stokes & $512^2$ & $22.2$ & -- \\
\bottomrule
\end{tabular}
\end{table}

\appendixsection{Training and Sampling Settings}
\label{app:implementation}

\appendixsubsection{Latent Model Initialization and Training}
\label{app:spectral}
\label{app:warmstart}
\label{app:network}

The latent field model is initialized from a principal-component approximation of the training fields. The encoder, branch network and trunk network each combine a linear mapping with a multilayer perceptron. The linear mappings are initialized from the leading principal components, with the encoder producing whitened latent scores and the branch recovering their original scales. The decoder initialization also incorporates the empirical mean field as a separate trunk direction. The nonlinear components are initialized at zero output, so training begins from the linear reconstruction.

All parameters are then trained jointly with Adam under a cosine learning-rate schedule. The autoencoder minimizes the mean-squared reconstruction error at $64$ grid points drawn at random in each step, so the decoder is trained as a function of the coordinate; the trunk acts on Fourier features of the coordinates. Once trained, the encoder produces the latent codes used for flow matching, which are standardized coordinate-wise. Their empirical covariance defines $\Sigma_z$, which is used in the local Gaussian prior and MALA preconditioner. The latent flow is a multilayer perceptron with a sinusoidal time embedding, trained using the independent coupling and the objective in \eqref{eq:flow-matching-loss}. The model dimensions, optimization settings and training budgets are summarized in \Cref{tab:params}.

\appendixsubsection{Local MALA Sampling}
\label{app:mala}

At each annealing level, LAPS approximates the local posterior in \eqref{eq:local-target} using $L$ preconditioned MALA steps. The preconditioner follows the Gauss--Newton form used in likelihood-informed and geometric samplers \citep{girolami2011riemann,cui2016dimension}:
\begin{equation}
    P_k
    =
    \left(
        \sigma^{-2}J^\top J
        +
        \lambda_k^{-2}\Sigma_z^{-1}
    \right)^{-1},
    \label{eq:gn-precond}
\end{equation}
where $J$ is the Jacobian of the latent observation map $G_\psi$ evaluated at $\vzero$. The Jacobian is computed once for each observation set, and the preconditioner is updated according to the local prior width $\lambda_k$.

Each chain is initialized from the local Gaussian prior:
\begin{equation}
    \vz^{(0)}
    =
    \alpha_k\bar{\vz}_k
    +
    \lambda_k L_z\veps_0,
    \qquad
    \Sigma_z=L_zL_z^\top,
    \qquad
    \veps_0\sim\N(\vzero,I_d),
    \label{eq:chain-start}
\end{equation}
where $L_z$ is the Cholesky factor of $\Sigma_z$. A diagonal jitter of $10^{-6}$ is added to $\Sigma_z$ and to $P_k$ before their Cholesky factorizations, so both are positive definite in the sampler; \Cref{thm:covariance-preservation} itself needs only a positive semidefinite $\Sigma_z$. The Metropolis--Hastings acceptance probability accounts for both the local posterior density and the forward and reverse Gaussian proposal densities. With a fixed positive-definite preconditioner and step size, the resulting transition leaves the local posterior invariant, although a finite number of MALA steps does not guarantee convergence.

The step sizes are tuned separately at each annealing level, toward an acceptance rate of $0.574$ on two of the held-out observation scenarios, and remain fixed during testing. The number of MALA steps and the other sampling parameters are given in \Cref{tab:params}.

\appendixsubsection{Annealing and Numerical Settings}
\label{app:schedule}
\label{app:settings}
\label{app:parameters}

The annealing schedule is uniform, with $t_k=k/K$ for $k=0,\ldots,K$. The residual-noise width follows \eqref{eq:bridge-scale}, and the learned flow is integrated using the explicit Euler method. The reported posterior experiments use $S=20$ transport steps, with the annealing levels $K$ and MALA steps $L$ specified for each benchmark in \Cref{tab:params}.

For numerical stability, the VP noise coefficient is bounded below by
\begin{equation}
    \lambda_k
    \leftarrow
    \max\{\lambda_k,\lambda_{\min}\},
    \qquad
    \lambda_{\min}=0.05,
\end{equation}
while $\alpha_k$ remains unchanged. The same width floor is applied to the original-coefficient variant in the ablation study. When the floor is active, the implemented coefficients need not satisfy $\alpha_k^2+\lambda_k^2=1$. With the uniform schedule this occurs only at the last level for $K=40$ and at the last three levels for $K=80$, where $\alpha_k^2+\lambda_k^2$ exceeds one by less than $\lambda_{\min}^2=0.0025$. The resulting covariance deviation is characterized by the general identity in \Cref{app:covariance-preservation}.

The numerical configurations used throughout the experiments are collected in \Cref{tab:params}. The data and observation settings are specified separately in \Cref{app:benchmark-construction}. The bimodal latent flow is wider and trained for longer than the others. Its configuration was fixed from likelihood-free diagnostics alone---the valley mass of decoded prior draws compared with that of the training fields, and the mode proportion under the annealing recursion run without a likelihood---so no test observation enters the choice.

\begin{table}[tbp]
\centering
\small
\setlength{\tabcolsep}{4pt}
\caption{Model, training and posterior sampling settings for the reported LAPS experiments. Flow widths are those of the hidden layers. The autoencoder validation error is the relative $L^2$ reconstruction error on the validation fields, averaged over the five training seeds and given separately for Mat\'ern and Gibbs.}
\label{tab:params}
\resizebox{\linewidth}{!}{
\begin{tabular}{lccc}
\toprule
Setting & Mat\'ern / Gibbs & Bimodal & Navier--Stokes \\
\midrule
\multicolumn{4}{l}{\textit{Latent model}} \\
Latent dimension $d$
    & $32$ & $32$ & $128$ \\
Decoder rank $r$
    & $64$ & $64$ & $256$ \\
Spatial Fourier frequencies
    & $64$ & $64$ & $64$ \\
Flow widths
    & $256\times2$ & $512\times4$ & $64,128,256,512,1024$ \\
\midrule
\multicolumn{4}{l}{\textit{Training}} \\
Autoencoder steps
    & $15{,}000$ & $15{,}000$ & $15{,}000$ \\
Flow training steps
    & $15{,}000$ & $100{,}000$ & $60{,}000$ \\
Batch size
    & $256$ & $256$ & $1024$ \\
Learning rate
    & $10^{-3}$ & $10^{-3}$ & $10^{-3}$ \\
Training seeds
    & $5$ & $5$ & $5$ \\
Autoencoder validation error
    & $0.012$ / $0.0007$ & $0.0009$ & $0.0030$ \\
\midrule
\multicolumn{4}{l}{\textit{Posterior sampling}} \\
Annealing levels $K$
    & $40$ & $40$ & $80$ \\
Euler steps $S$
    & $20$ & $20$ & $20$ \\
MALA steps $L$
    & $50$ & $50$ & $100$ \\
Posterior draws $N$
    & $128$ & $128$ & $128$ \\
Residual noise $\sigma_{\min}$
    & $10^{-3}$ & $10^{-3}$ & $10^{-3}$ \\
Width floor $\lambda_{\min}$
    & $0.05$ & $0.05$ & $0.05$ \\
\bottomrule
\end{tabular}
}
\end{table}

For the FAPS comparison, we use the published sampler configuration with $K=40$, $S=20$ and $L=50$, together with a rank-$32$ covariance estimated from $256$ prior samples. Released prior models are used for the Gaussian benchmarks, while the bimodal and Navier--Stokes priors are trained using the authors' code and the corresponding training fields. Four additional prior models are trained for each benchmark to obtain five training seeds. These configurations are kept fixed in the baseline comparison, whose reproduction results and evaluation conditions are described in \Cref{app:reproduction}.

\appendixsection{Further Evaluation of Posterior Sampling}
\label{app:supplementary}

\appendixsubsection{Annealed Sampling and Direct MCMC}
\label{app:direct-mcmc}

The schedule ablation in \Cref{sec:exp-ablation} compares multilevel annealing with a single-level Gaussian replacement. Since the latter changes both the annealing schedule and the effective prior, we further compare LAPS with direct MCMC methods that retain the same learned latent prior.

Direct MCMC operates in the Gaussian reference space of the learned flow. Using the reference variable $\vz_0$, the target density is
\begin{equation}
    p(\vz_0\mid\vy)
    \propto
    \exp\!\left[
        -\Phi_\psi\bigl(T^\theta_{0,1}(\vz_0);\vy\bigr)
    \right]
    \N(\vz_0;\vzero,I_d),
\end{equation}
where $T^\theta_{0,1}$ maps the Gaussian reference to the learned latent prior. Unlike the initial annealing distribution, this target incorporates the observation likelihood.

We consider preconditioned Crank--Nicolson (pCN) \citep{cotter2013mcmc} and MALA. Each scenario uses $128$ independent chains initialized from the standard Gaussian reference, with $S=20$ Euler steps for flow transport. The final reference states are transported to the latent space and decoded. The pCN proposal is tuned on held-out scenarios to an acceptance rate of $0.234$ and then fixed. For MALA, the Gauss--Newton preconditioner is
\begin{equation}
    P_0
    =
    \left(
        \sigma^{-2}J_0^\top J_0+I_d
    \right)^{-1},
\end{equation}
where $J_0$ is the Jacobian of $G_\psi\circ T^\theta_{0,1}$ evaluated at $\vzero$. The MALA step size is adapted during the first half of each chain and fixed thereafter.

The iteration budgets are selected according to the recorded wall-clock cost of LAPS. The Gaussian benchmarks use $1000$ pCN or $450$ MALA steps, while Navier--Stokes uses $2800$ or $1250$ steps. On the bimodal benchmark, whose flow network is larger, the corresponding budgets are $725$ and $360$ steps. Direct MCMC therefore receives matched or more generous computational budgets.

\begin{table}[t]
\centering
\small
\setlength{\tabcolsep}{3pt}
\caption{Comparison of LAPS and direct MCMC under finite computational budgets. Results are means $\pm$ standard deviations over five training seeds and $100$ test scenarios, with $128$ posterior draws per scenario. Direct MCMC uses the same learned latent prior and approximately matched or more generous wall-clock budgets.}
\label{tab:direct-mcmc}
\resizebox{\linewidth}{!}{

\begin{tabular}{lcccccccc}
\toprule
 & \multicolumn{2}{c}{Mat\'ern} & Gibbs & \multicolumn{2}{c}{Bimodal} & \multicolumn{3}{c}{Navier--Stokes} \\
\cmidrule(lr){2-3}\cmidrule(lr){4-4}\cmidrule(lr){5-6}\cmidrule(lr){7-9}
Sampler & SWD $\downarrow$ & MAE$_s$ $\downarrow$ & SWD $\downarrow$ & SWD $\downarrow$ & ModeW-MAE $\downarrow$ & rel.\ $L^2$ $\downarrow$ & CE $\downarrow$ & SSR $\to1$ \\
\midrule
LAPS & $0.0614_{\pm 0.0007}$ & $0.0133_{\pm 0.0005}$ & $0.0762_{\pm 0.0006}$ & $0.1499_{\pm 0.0090}$ & $0.0661_{\pm 0.0071}$ & $0.0369_{\pm 0.0007}$ & $0.0047_{\pm 0.0011}$ & $0.989_{\pm 0.012}$ \\
direct pCN & $0.0617_{\pm 0.0012}$ & $0.0159_{\pm 0.0006}$ & $0.0756_{\pm 0.0014}$ & $0.1401_{\pm 0.0054}$ & $0.0568_{\pm 0.0043}$ & $0.0399_{\pm 0.0004}$ & $0.1185_{\pm 0.0024}$ & $1.689_{\pm 0.010}$ \\
direct MALA & $0.1302_{\pm 0.0336}$ & $0.0571_{\pm 0.0252}$ & $0.1207_{\pm 0.0262}$ & $0.2710_{\pm 0.0205}$ & $0.1489_{\pm 0.0070}$ & $0.2307_{\pm 0.0027}$ & $0.1505_{\pm 0.0051}$ & $1.571_{\pm 0.018}$ \\
\bottomrule
\end{tabular}
}
\end{table}
\FloatBarrier

As shown in \Cref{tab:direct-mcmc}, direct pCN achieves SWD comparable to LAPS on the Gaussian benchmarks and slightly lower SWD on the bimodal benchmark. On Navier--Stokes, LAPS provides substantially better marginal calibration at the tested budget. The comparison characterizes sampling at a finite budget and does not establish superiority over converged direct MCMC.

\appendixsubsection{Posterior Mode-Weight Accuracy}
\label{app:mode-weights}

The bimodal results in \Cref{sec:exp-bimodal} establish the recovery of two separated posterior configurations. To further assess the distribution of probability mass between them, we compare the sampled mode weights with the exact mixture weights across all $100$ test scenarios. The sampled weights are estimated using the fixed template assignment defined in \Cref{app:metrics}.

\begin{figure}[t]
\centering
\includegraphics[width=0.8\linewidth]{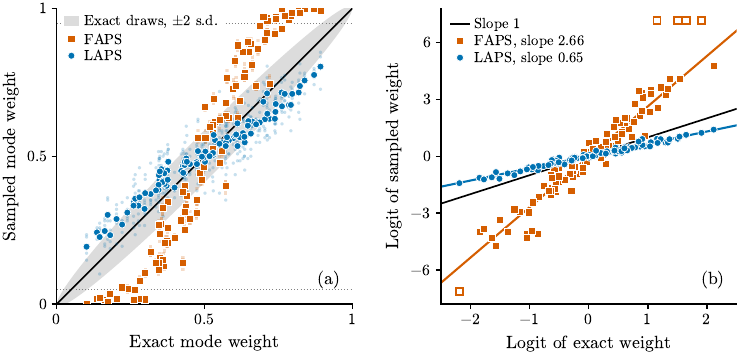}
\caption{Sampled versus exact posterior mode weights on the bimodal benchmark across $100$ test scenarios and five training seeds. (a) Linear scale: faint points represent individual training seeds and solid markers their means. The gray band shows $\pm2$ standard deviations for $128$ independent exact-posterior draws, and dotted lines indicate the $5\%$ and $95\%$ thresholds. (b) Logit scale: sampled weights are pooled across five seeds ($640$ draws per scenario), with solid lines showing least-squares fits. Open markers denote scenarios in which all pooled draws belong to one configuration and are excluded from the fits.}
\label{fig:bimodal-weights}
\end{figure}

\Cref{fig:bimodal-weights} presents this comparison on linear and logit scales. LAPS follows the variation in the exact weights, with most estimates lying within the finite-sample variation of independent exact-posterior draws. Its logit-scale slope is approximately $0.65$, ranging from $0.64$ to $0.66$ across the five training seeds, indicating a systematic bias toward more balanced mixtures. FAPS exhibits the opposite tendency, with a slope of $2.66$ among scenarios in which both configurations receive samples. Its sampled weights also fall below $5\%$ or above $95\%$ in $27\%$ of the scenarios.

Recovering both modes is thus distinct from weighting them correctly: LAPS retains both configurations in every scenario, but its mode weights are biased toward one half.

\appendixsubsection{Bridge Coefficients: All Metrics}
\label{app:bridge-full}

\Cref{tab:ablation-bridge-full} extends \Cref{tab:ablation-bridge} to every scored metric, under the protocol of \Cref{sec:exp-ablation}. On the three exact-reference benchmarks every distance to the exact posterior improves under VP in every training seed. On Navier--Stokes the calibration measures improve, CRPS increases slightly and the relative $L^2$ error is unchanged within the paired interval.

\begin{table}[tbp]
\centering
\small
\caption{Original versus VP bridge coefficients, all scored metrics. Conventions are those of \Cref{tab:ablation-bridge}.}
\label{tab:ablation-bridge-full}
\resizebox{\linewidth}{!}{

\begin{tabular}{llcccc}
\toprule
Benchmark & Metric & Original & VP & VP $-$ original [95\% CI] & Seeds \\
\midrule
Mat\'ern & SWD $\downarrow$ & $0.0809_{\pm 0.0011}$ & $0.0614_{\pm 0.0007}$ & $-0.0195$ \,[$-0.0204$,\,$-0.0186$] & 5/5 \\
 & MMD $\downarrow$ & $0.0268_{\pm 0.0017}$ & $0.0147_{\pm 0.0012}$ & $-0.0121$ \,[$-0.0144$,\,$-0.0098$] & 5/5 \\
 & $\mathrm{RMSE}_m$ $\downarrow$ & $0.0312_{\pm 0.0012}$ & $0.0266_{\pm 0.0011}$ & $-0.0046$ \,[$-0.0052$,\,$-0.0040$] & 5/5 \\
 & $\mathrm{MAE}_s$ $\downarrow$ & $0.0356_{\pm 0.0009}$ & $0.0133_{\pm 0.0005}$ & $-0.0224$ \,[$-0.0232$,\,$-0.0216$] & 5/5 \\
 & width $\to1$ & $1.108_{\pm 0.003}$ & $0.972_{\pm 0.002}$ & $-0.080$ \,[$-0.086$,\,$-0.074$]$^{\ast}$ & 5/5 \\
\addlinespace[0.25em]
Gibbs & SWD $\downarrow$ & $0.1008_{\pm 0.0009}$ & $0.0762_{\pm 0.0006}$ & $-0.0245$ \,[$-0.0255$,\,$-0.0236$] & 5/5 \\
 & MMD $\downarrow$ & $0.0389_{\pm 0.0021}$ & $0.0180_{\pm 0.0027}$ & $-0.0209$ \,[$-0.0226$,\,$-0.0191$] & 5/5 \\
 & $\mathrm{RMSE}_m$ $\downarrow$ & $0.0377_{\pm 0.0014}$ & $0.0317_{\pm 0.0015}$ & $-0.0060$ \,[$-0.0066$,\,$-0.0054$] & 5/5 \\
 & $\mathrm{MAE}_s$ $\downarrow$ & $0.0395_{\pm 0.0005}$ & $0.0147_{\pm 0.0004}$ & $-0.0248$ \,[$-0.0252$,\,$-0.0245$] & 5/5 \\
 & width $\to1$ & $1.090_{\pm 0.003}$ & $0.972_{\pm 0.002}$ & $-0.062$ \,[$-0.067$,\,$-0.056$]$^{\ast}$ & 5/5 \\
\addlinespace[0.25em]
Bimodal & SWD $\downarrow$ & $0.1658_{\pm 0.0033}$ & $0.1499_{\pm 0.0090}$ & $-0.0159$ \,[$-0.0268$,\,$-0.0050$] & 5/5 \\
 & MMD $\downarrow$ & $0.0694_{\pm 0.0030}$ & $0.0529_{\pm 0.0079}$ & $-0.0165$ \,[$-0.0258$,\,$-0.0073$] & 5/5 \\
 & $\mathrm{RMSE}_m$ $\downarrow$ & $0.0822_{\pm 0.0029}$ & $0.0678_{\pm 0.0056}$ & $-0.0144$ \,[$-0.0210$,\,$-0.0078$] & 5/5 \\
 & $\mathrm{MAE}_s$ $\downarrow$ & $0.0217_{\pm 0.0011}$ & $0.0171_{\pm 0.0007}$ & $-0.0046$ \,[$-0.0061$,\,$-0.0032$] & 5/5 \\
 & ModeW-MAE $\downarrow$ & $0.0782_{\pm 0.0038}$ & $0.0661_{\pm 0.0071}$ & $-0.0121$ \,[$-0.0209$,\,$-0.0032$] & 5/5 \\
 & ModeCov $\uparrow$ & $1.000_{\pm 0.000}$ & $1.000_{\pm 0.000}$ & $+0.000$ \,[$+0.000$,\,$+0.000$] & tie \\
 & Valley $\to0.04$ & $0.057_{\pm 0.002}$ & $0.045_{\pm 0.001}$ & $-0.012$ \,[$-0.014$,\,$-0.009$] & 5/5 \\
\addlinespace[0.25em]
Navier--Stokes & CRPS $\downarrow$ & $0.0156_{\pm 0.0003}$ & $0.0158_{\pm 0.0003}$ & $+0.0002$ \,[$+0.0001$,\,$+0.0003$] & 0/5 \\
 & rel.\ $L^2$ $\downarrow$ & $0.0367_{\pm 0.0008}$ & $0.0369_{\pm 0.0007}$ & $+0.0002$ \,[$-0.0001$,\,$+0.0004$] & 1/5 \\
 & CE $\downarrow$ & $0.0095_{\pm 0.0034}$ & $0.0047_{\pm 0.0011}$ & $-0.0048$ \,[$-0.0101$,\,$+0.0005$] & 4/5 \\
 & SSR $\to1$ & $1.057_{\pm 0.023}$ & $0.989_{\pm 0.012}$ & $-0.044$ \,[$-0.081$,\,$-0.007$]$^{\ast}$ & 5/5 \\
\bottomrule
\end{tabular}
}
\end{table}

\bibliographystyle{elsarticle-num-names}
\bibliography{laps}

\end{document}